\documentclass[letterpaper, 10 pt, conference]{ieeeconf}  % \usepackage{microtype}
\usepackage{times}
\usepackage{authblk}
\usepackage[compress,numbers]{natbib}
\usepackage{multicol}
\usepackage{graphicx,caption,subcaption} 
\usepackage[font={small}]{caption}

\usepackage{amsmath,amsfonts,amssymb}
\usepackage{algorithm,algpseudocode}
\usepackage{url}
\usepackage{tikz}
\usetikzlibrary{arrows}
\graphicspath{{figures/}}
\usepackage[capitalize]{cleveref}
\crefformat{equation}{(#2#1#3)}
\Crefformat{equation}{Equation~(#2#1#3)}
\Crefname{equation}{Equation}{Equations}
\crefformat{figure}{Fig.~#1}

\allowdisplaybreaks[2]
\renewcommand{\baselinestretch}{0.97}
\newtheorem{assumption}{Assumption}

\begin{document}

\title{\huge SLAM with Objects using a Nonparametric Pose Graph}
\author[1]{Beipeng Mu}
\author[1]{Shih-Yuan Liu}
\author[2]{Liam Paull}
\author[2]{John Leonard}
\author[1]{Jonathan P.\ How}
\affil[1]{Laboratory for Information and Decision Systems}
\affil[2]{Computer Science and Artificial Intelligence Laboratory}
\affil[ ]{Massachusetts Institute of Technology, \{mubp, syliu, lpaull, jhow, jleonard\}@mit.edu}
\maketitle

\begin{abstract}
Mapping and self-localization in unknown environments are fundamental capabilities in many robotic applications.
These tasks typically involve the identification of objects as unique features or landmarks, which requires the objects both to be detected and then assigned a unique identifier  that can be maintained when viewed from different perspectives and in different images. 
The \textit{data association} and \textit{simultaneous localization and mapping} (SLAM) problems are, individually, well-studied in the literature. But these two problems are inherently tightly coupled, and that has not been well-addressed.
Without accurate SLAM, possible data associations are combinatorial and become intractable easily.
Without accurate data association, the error of SLAM algorithms diverge easily.
This paper proposes a novel nonparametric pose graph that models  data association and SLAM in a single framework.
An algorithm is further introduced to alternate between inferring data association and performing SLAM.
Experimental results show that our approach has the new capability of associating object detections and localizing objects at the same time, leading to significantly better performance on both the data association and SLAM problems than achieved by considering only one and ignoring imperfections in the other.
\footnote{software is available at https://github.com/BeipengMu/objectSLAM.git}
\footnote{video is available at https://youtu.be/YANUWdVLJD4}
\end{abstract}

\section{Introduction}
In many robotics applications, such as disaster relief, planetary exploration, and surveillance, robots are required to autonomously explore unknown spaces without an accurate prior map or a global position reference (e.g. GPS).
A fundamental challenge faced by the robot is to effectively localize itself using only the information extracted from the environment.
For example, the capability of recognizing instances of objects and associating them with unique identifiers will enable the robot to build maps of the environment and localize itself within.
The problem of constructing a global map and localizing the robot within is referred as simultaneously localization and mapping (SLAM).

SLAM with various representations of the world and different sensors has been thoroughly studied in the literature.
Occupancy grid map with LiDAR or laser range finders is among the early successes that dates back to the 1980s \cite{occupancy_grid, Elfes_1989, Thrun_AR_2003, prob_robotics}.
In occupancy based approaches, the world is represented by 2D/3D grids composed of free spaces and occupied spaces. New scans from the
LiDAR or laser range finders are compared and matched with previous scans to incrementally build such maps.
This simplified representation of the world facilitates efficient computation, and thus real-time performance can be achieved on relatively large scenes with a single CPU. However, 
The successful matching of two scans relies on geometric features such as corners.
In places that lack such features, like long hallways, SLAM using occupancy grid maps tends to fail. 
In recent years, SLAM with 3D dense mapping and RGB-D cameras has become more and more popular \cite{Whelan-RSS-15, Keller_dense3d, Newcombe_kinectfusion}.
This line of work is able to utilize both the geometric information from depth cameras and the color information from RGB cameras to reconstruct environments in dense 3D maps.
Incoming depth and color images are converted into volumes or deformation surfaces~\cite{Whelan-RSS-15}, then matched with previously constructed volumes or surfaces to incrementally build the map.
3D dense maps provide photographic details of the environment with millions of volumes or surfaces.
However, they rely heavily on parallel computation enabled by GPUs, and do not scale very well.

\begin{figure}[t]
\centering
\includegraphics [width=0.8\columnwidth]{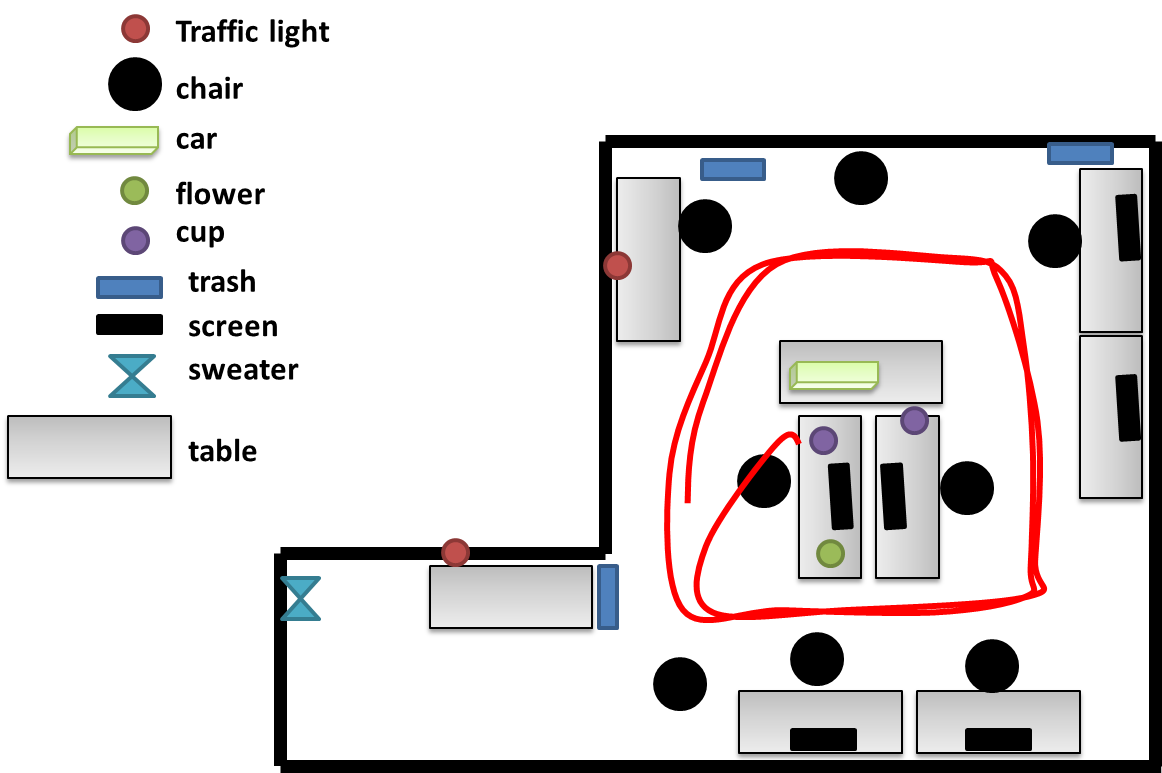}  
	\caption{\small In object SLAM, each object class has multiple instances, data association (associate detect objects to unique object idnetifiers) is ambiguous. Data association and SLAM (localize objects) are inherently coupled: good data association guarantees the convergence of SLAM, and good SLAM solution gives good initialization of data association.}\label{fig:office_floormap}
\end{figure} 

A factor graph is a different representation of the SLAM problem \cite{Mahon_IEEETRO_2008, Graham15_robustslam, Rosen12_isam, Kaess_IJRR_2012}. 
Instead of using small units, such as grids, volumes, or surfaces, to represent the space, a factor graph encodes the poses of the robot and the observed landmarks along the trajectory.
In a factor graph, each factor represents a constraint on the relative poses either between two consecutive robot poses or between a robot pose and a landmark.
The robot poses and landmark positions are modeled as random variables, and they are optimized by maximizing the joint likelihood.
Mechanisms can be designed such that new factors are only added when there are significant pose updates or new object measurements to facilitate concise representation. As a result, 
Factor graph SLAM scales much better than SLAM with occupancy grid maps or 3D dense maps.
However, the convergence of factor graph SLAM algorithms relies heavily on correct data association of the landmarks.
Even a single false association can cause the algorithm to diverge \cite{montemerlo2003simultaneous, Graham_robust}.

The focus on of this work is on SLAM in unknown environment by recognizing objects and utilizing their positions (object SLAM).
A factor graph is the natural representation, as objects can be easily represented as landmarks.
A map represented by objects is desirable, as objects are very rich in semantic meanings.
By using objects, robots can interact with other agents and perform tasks at semantic level, such as searching for people in a forest, grasping objects, and detecting moving cars on streets. 
Object SLAM requires the robot to be able to detect objects, generate measurements, and associate these measurements to unique identifiers.
In this work, object detection refers to the problem of identifying the occurrence of objects of some predefined object classes within an image. 
An object measurement is a 3D location of the detected object with respect to the robot pose.
Data association refers to the problem of associating object measurements to unique identifiers across images.
The problem of object detection has been an important topic in the computer vision community.
Deep learning approaches have achieved significant success on object detections within individual images \cite{Everingham10, girshick2014rich, Szegedy_2013_DL, ILSVRC15, Erhan_2014_DL}.
These approaches also have the ability to generalize: once a detector is trained to recognize a object class, such as chairs, the detector can detect different instances of the same class even in different shape, color, and background settings.
Some recent work on Region-based Convolutional Neural Networks \cite{girshick2014rich, renNIPS15fasterrcnn} gained significant success on training deep learning models to detect multiple objects instances within a single image.
However, object detections only suggest the existence of objects of certain predefined object classes in an image, but provide no data association between images.
given that an object of a certain class is detected in two images, the object detector provides no information on whether or not the detected objects in the two images are the same object.
This is problematic for SLAM especially when there are multiple objects of the same object class in an environment. 
How reliably SLAM can be achieved using only these ambiguous object detections remains an open question.
As illustrated in \Cref{fig:office_floormap}, there are multiple instances of the same object class, such as chairs. The robot would need to establish the data association of object detections across images from different views. Note that data association and SLAM are inherently coupled problems: good data association guarantees the convergence of SLAM algorithms, and good SLAM solution gives good initialization of data association.

This paper proposes a novel world representation, the nonparametric pose graph, to jointly perform data association and SLAM.
In the proposed model, factor graphs are used to localize robots and objects, while a Dirichlet process (DP) -- a nonparametric model prior -- is used to associate detections to unique object identifiers in the scene.
The inference of the data associations and the optimization of the the robot and object poses are performed alternatively in this algorithm.
This coupled framework achieves better performance for both data association and SLAM.

The contributions of this work are:
\begin{itemize}
\item Creating a nonparametric pose graph model that couples data association and SLAM. 
\item Proposing an algorithm that jointly infers data associations and optimizes robot poses/object locations over nonparametric pose graphs.
\item Developing an approach to generate object measurements from RGB and depth images in 3D space via deep learning object detection.
\item Demonstrating the performance of the proposed approach via both simulated and real-world data.
\end{itemize}

\section{Related Work}
Data association of objects and SLAM are typically solved as decoupled problems in the literature.
Pillai and Leonard \cite{Pillai-RSS-15} showed that when the SLAM solution is known, and thus there is no uncertainty in robot poses, robot poses provide good prior information about object locations and can achieve better recalls than frame by frame detections. 
Song et~al. \cite{Song_humanobj} used a SLAM solver to build a 3D map of a room, and then fixed the map and manually labeled objects in the room.
On the other hand, object detection can improve localization as well.
Atanasov et al. \cite{Atanasov-RSS-14} pre-mapped doors and chairs as landmarks.
During the navigation stage, these pre-mapped objects are detected online and their location information is used to localize the robot. 

However, in the scenario considered here, neither data association of objects nor robot poses are perfectly known.
The algorithm must associate objects detections and perform SLAM simultaneously.
Algorithms that solve object detection and SLAM jointly can be categorized into front-end approaches and back-end approaches.
\subsection{Front-end Data Association}
In front-end approaches, objects detected in  new images are compared with previous images.
If matches between new and old images are found, then corresponding objects are associated to the same unique identifier.
These matches are typically reliable as the disparity between two consecutive images are usually small.
When the robot come back to a previously visited place after traversing a long distance, costly global optimization must be performed to achieve global loop closures.
These data associations by front-end procedures are taken as reliable and true, and then passed to a SLAM solver \cite{slam++object, Civera_semanticslam}.
SLAM++ \cite{slam++object} is one such front-end approach.
Full 3D scans of chairs and tables are created and used as templates. When new point clouds are observed during testing, they are matched to pre-built templates.
Successfully matched detections are often of high credibility. A SLAM solver is then run on these reliable detections to optimize object locations and camera poses.
In semantic SLAM \cite{Civera_semanticslam}, Civera et al. created a library of six objects, used SURF features to detect these objects, and then ran an EKF to simultaneously localize robot and map objects.

In this work, instead of creating exact templates for objects, deep learning is used to detect objects in the environment.
Deep learning generalizes much better than template-based approaches.
It can leverage open source software (millions of images already exist online to create models), scales easily to hundreds of object classes instead of a handful of pre-tuned templates, and does not require the objects in the scene to be exactly the same as the templates.
However, the detections have significant ratio of false positives and partial occlusions, thus are very challenging for front-end algorithms to produce reliable data associations.

\subsection{Back-end Robust SLAM}
Robust SLAM is a line of research that explicitly use back-end approaches to deal with outliers in the data \cite{Sunderhauf12_robustslam,Graham15_robustslam,Latif13_robust}.
In robust SLAM, most of the object measurements are already correctly associated to unique identifiers. 
when some measurement is incorrectly associated, it will be inconsistent with other object measurements of the same identifier. %Robust SLAM tries to identify these inconsistent measurements and eliminate them.
%When the remaining measurements are consistent with each other, standard SLAM solvers can be used to optimize poses.
%
%It was demonstrated in \cite{Graham15_robustslam} that directly using measurements with wrong identifiers leads to divergence of the pose graph SLAM solutions.
Robust SLAM instead maximizes a set of measurements that are consistent with each other in both identifiers and predicted locations.
Only the consistent measurements are plugged into a SLAM solver to recover the robot poses and landmark locations. 

By nature robust SLAM relies on the assumption that inlier measurements with unique identifier associations are the majority compared to outlier measurements. Under this assumption, eliminating outliers can still give good SLAM results.
However, in object SLAM, it is often the case that there are multiple instances of the same object class.
If all object measurements with same class are associated to the same identifier, different object instances will always give inconsistent measurements.
In other words, in object SLAM, outliers are pervasive.
If only one set of consistent measurements for each object class is kept, the algorithm will eliminate the majority of the data and fail  to identify any repetitive instances of the same class.

The algorithm presented in this paper is a back-end approach where there are multiple instances of the same object class. The data association of object measurements to unique identifiers are considered unknown and must be established while doing SLAM. 
We exploit the coupling between data association and SLAM, jointly optimize both, and achieve better performance on both.

\section{Object Measurements via Deep Learning}
This section sets up the approach to generate object measurements via deep learning. The limitations of such an approach are discussed at the end of the section, which highlights the necessity of back-end data association and SLAM algorithms.

\subsection{Deep Learning Based Object Detection}
Object detection refers to the problem of identifying the existence of objects of certain classes and find bounding boxes for them in single images. Object detection in the past decade was mainly based on the use of SIFT and HOG features. Although researchers have developed algorithms that demonstrated good performance for single class object detection (e.g. pedestrians), the multi-class object detection problem remains difficult. In particular, prior to 2012, the state-of-the-art method (deformable part models) achieved 33.4\% accuracy on the PASCAL VOC 2010 dataset \cite{girshick2014rich}, which contains 20 object classes. 

Region-based convolutional neural network (R-CNN) \cite{girshick2014rich} was among the first works on object detection using a CNN.
This algorithm uses the selective search \cite{uijlings2013selective} algorithm to generate bounding box proposals, and then crops an image patch using each proposal.
The image patches are subsequently scaled and run through a CNN model for object detection.
This approach achieved 53.7\% accuracy on the PASCAL VOC dataset. 
R-CNN is extremely slow (13 seconds per image) because all patches need to run through the CNN individually.

In Faster R-CNN \cite{renNIPS15fasterrcnn}, Ren et. al. ran the full image through the CNN only once, and they only use features in top layer in each bounding box patch for object detection. They further proposed a region proposal network (RPN) that learns how to generate bounding box proposals by looking at the top layer features. This new algorithm achieves 76\% accuracy and 
an average speed of 100 milliseconds per image.

Faster R-CNN \cite{renNIPS15fasterrcnn} uses the VOC dataset for training. Most of the object classes in the VOC dataset \cite{Everingham10} are rare in urban or indoor settings, such as cows, horses, sheeps, airplanes, and boats. 
Our work trained a faster R-CNN model on the ImageNet 2014 dataset \cite{ILSVRC15}, which contain categories that are more relevant to indoor/urban settings, including \textit{cars, motorcycles, bicycles, traffic lights, televisions, chairs, flowerpots, cups, and keyboards}.
Note that this framework can be easily modified to parse out any other subset of classes from the ImageNet dataset that are relevant to the specific applications.

\begin{figure}[ht!]
\centering
\begin{subfigure}[b]{0.225\textwidth}
	\centering
	\includegraphics [clip = true, trim=20 0 0 0, width=0.9\columnwidth, angle = 0]{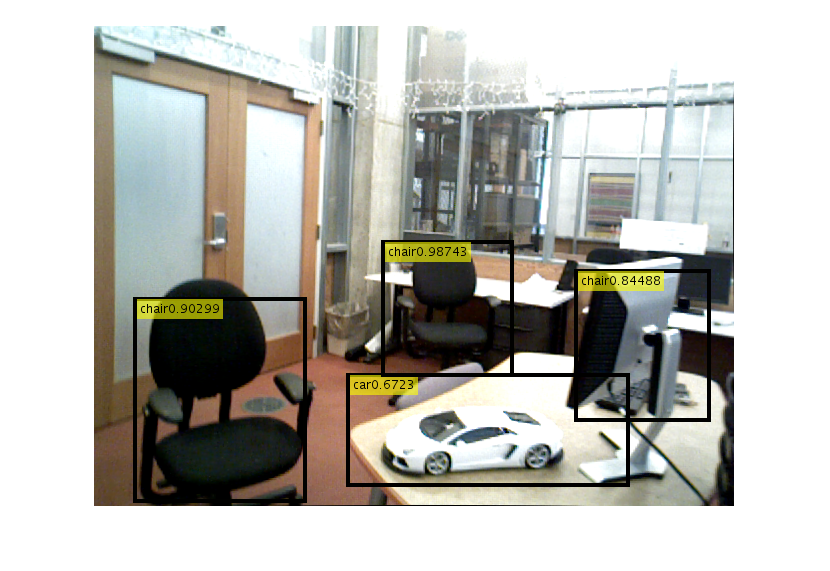}  
	\caption{Object detection with RGB image.} 
	\label{fig::obj_rgb}
\end{subfigure}
\begin{subfigure}[b]{0.225\textwidth}
	\centering
	\includegraphics [clip = true, trim=0 0 0 0, width=0.9\columnwidth, angle = 0]{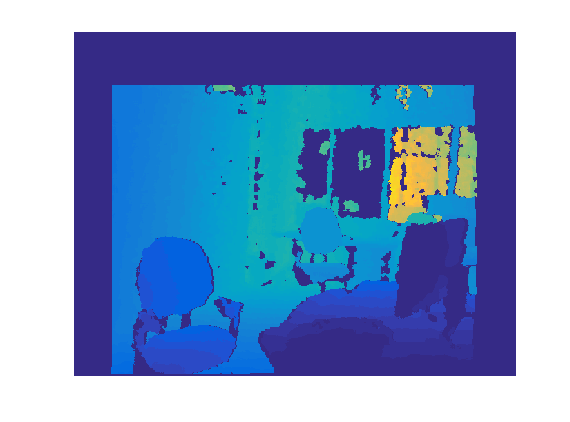} 
	\caption{The depth image corresponding to the RGB image.} 
	\label{fig::obj_depth}
\end{subfigure} \\
\begin{subfigure}[b]{0.3\textwidth}
	\centering
	\includegraphics [clip = true, trim=80 80 50 0, width=0.7\columnwidth, angle = 0]{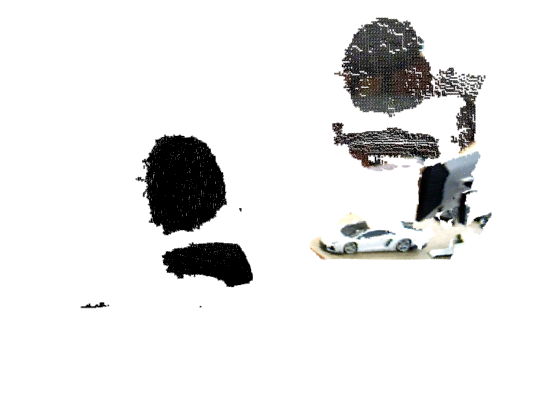}  
	\caption{Object localization in 3D space} 
	\label{fig::obj_cloud}
\end{subfigure}
\caption{\small Deep learning based object detection}
\end{figure}

\subsection{Object Measurements}
An object measurement refers to a labeled 3D location with respect to the robot pose. To generate such measurements, location information relative to the robot is required in addition to object detection. In this paper, this is done by inquiring the corresponding pixels in the depth images: (1) Crop bounding boxes in the depth image in correspondence with the RGB bounding box. 
(2) Filter out background pixels that are too far away. 
(3) Generate point cloud from RGB and depth pairs. 
(4) Compute the centroid of the point cloud as center of the object.

\Cref{fig::obj_rgb} shows the detected object with faster R-CNN from a single image of an office environment. \Cref{fig::obj_depth} shows the corresponding depth image, and \cref{fig::obj_cloud} shows the four point clouds for the four detected objects in 3D space.

It is clear from \cref{fig::obj_rgb} that object SLAM with deep learning object detection has two major challenges.
First, there are multiple instances of the object class, such as ``chair'' in \cref{fig::obj_rgb}.
Without correct data association, it is hard to distinguish different object instances. Standard pose-graph SLAM algorithms can only optimize poses with exact data association, such as g2o\cite{g2o}, isam\cite{Rosen12_isam}, gtsam\cite{gtsam}.
The second challenge is high false positive rates. 
As the \texttt{chair} detected in \cref{fig::obj_rgb}, deep learning algorithms report objects now and then when there are actually none.
Blindly using these unfiltered detections in standard SLAM algorithms will lead to ``non-exist" nodes and cause loop closure failures. 

Notice that the centroid is used as the center of objects in this case. When objects are looked at from different views, and be partially occluded, centroids would not be a consistent measure of the object locations. In our experience, the error could be 10-20cm. However, we will show that in office settings, our algorithm still converges even under  significant occlusion and view point noise.

\section{Pose Graph Background}
This section sets up the background on graphical models used in SLAM problems. The next section will extend the current pose graph to a novel nonparametric pose graph and introduce an algorithm to perform inference on it.

\subsection{Factor Graphs}
\begin{figure}[t!]
	\centering
\resizebox{.75\columnwidth}{!}{%	
	\begin{tikzpicture}
	[hidden/.style={circle, draw, minimum size = 0.8cm},
	factor/.style={rectangle, draw, minimum size = 0.2cm}]
	% hidden variables
	\node[hidden] (x1) at (-3,-1.5) {$X_1$};
	\node[hidden] (x2) at (0,0) {$X_2$};
	\node[hidden] (x3) at (0,-3) {$X_3$};
	\node[hidden] (x4) at (3,-1.5) {$X_4$};
	\node[hidden] (x5) at (6,-1.5) {$X_5$}; 

	\node[factor] (f12) at (-1.5,-0.75) {$\psi_{12}$};
	\node[factor] (f13) at (-1.5,-2.25) {$\psi_{13}$};
	\node[factor] (f23) at (0,-1.5) {$\psi_{23}$};
	\node[factor] (f34) at (1.5,-2.25) {$\psi_{34}$};
	\node[factor] (f45) at (4.5,-1.5) {$\psi_{45}$};
	
	% edges 
	\draw[blue,thick] (x1) --(f12); \draw[blue,thick] (f12) --(x2);
	\draw[blue,thick] (x1) --(f13); \draw[blue,thick] (f13) --(x3);
	\draw[blue,thick] (x2) --(f23); \draw[blue,thick] (f23) --(x3);	
	\draw[blue,thick] (x3) --(f34); \draw[blue,thick] (f34) --(x4);
	\draw[blue,thick] (x4) --(f45); \draw[blue,thick] (f45) --(x5);
	\end{tikzpicture}}
	\caption[Factor graph]{Factor graph. Squares denote factors: $\psi_{12}(X_1,X_2)$, $\psi_{23}(X_2,X_3)$, $\psi_{13}(X_1,X_3)$, $\psi_{34}(X_3,X_4)$, and $\psi_{45}(X_4,X_5)$.}\label{fig:factorgraph}
\end{figure}
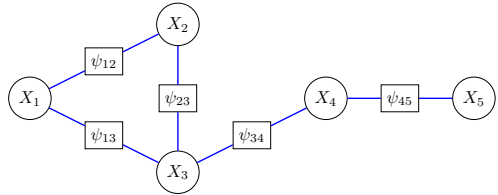

A graphical model is a probabilistic model that expresses the conditional dependence structure between random variables. Graphical models use a graph-based representation to encode a complete distribution over a high-dimensional space \cite{Bishop07_PRML}. Commonly used graphical models include Bayesian networks, Markov Random Fields, and factor graphs. 
\cref{fig:factorgraph} gives an example of such a factor graph. Circles represent random variables and squares represent factors.
A factor graph is a graphical model widely used in SLAM problems. Denote $\mathbf X=\{X_1,\cdots,X_n\}$ as the random variables. Denote $\psi_a(X_{\{a\}})$ as a factor among random variable in set $\{a\}$. The the joint probability can be expressed as a product of factors:  
\begin{equation}
p(\mathbf X=x)  \propto \prod_{a\in \mathcal{A}} \psi_a \left(X_{\{a\}}=x_{\{a\}} \right)
\end{equation}
where $\mathcal{A}$ is the set of all factors.
Each factor $\psi_a(x_{\{a\}})$ maps the values of random variables to a strictly positive real number representing the likelihood of the variables.  It also represents probabilistic dependences among the variables in the factor.

Factor graph is an efficient representation in that it captures sparsity among variables: two variables are independent of each other given all other variables if and only if they do not belong to the same factor.
The log likelihood, $\log p(x)$, can be written in a sum of factors:
$
\log p(x)  \propto \sum_{a\in \mathcal{A}} \phi_a \left(x_{\{a\}} \right)
$,
where $\mathcal{A}$ is the set of all factors. Each factor $\phi_a(x_{\{a\}})$ maps the values of random variables to a strictly positive real number representing the log likelihood of the variables.
With graphical models, there exist fast algorithms to compute statistical properties such as marginalization, expectation, maximum likelihood \cite{Bishop07_PRML}.

\subsection{Factor Graph for SLAM}
Factor graphs have gained a lot of success and popularity for SLAM problems \cite{Rosen12_isam, gtsam} due to their efficiency. 
Fist assume that there exist static landmarks that the robot can identify to localize itself.
\begin{assumption}
There exists a library of static landmarks to localize the robot in the environment. The number and locations of these landmarks is not known a priori.
\end{assumption}

With the landmark assumption, when moving in the environment, the robot can obtain measurements of these landmarks. Given a dataset, the robot trajectory is represented as a discrete sequence of poses.
Denote $T$ as the total number of time steps, and denote $\mathbf X_{0:T} = \{X_0 , \cdots , X_T \}$ as the robot's trajectory from the start to the end.
Each robot pose consists of a position and an orientation. Denote $SE(2)$ as the space of 2D poses and $SE(3)$ as the space of 3D poses. Then $X_t \in SE(2)$ for 2D cases and $X_t \in SE(3)$ in 3D cases.
In GPS-denied environments these poses are not directly observable.
However, the robot can always measure the incremental change between two
sequential poses via an IMU or wheel encoder, which is referred to as odometry.
Denote $o_t$ as the odometry measurement between pose $x_t$ and pose $x_{t-1}$. Under the standard assumption that $o_t$ is corrupted by additive Gaussian noise, the odometry measurement at time $t$ can be represented as:
\begin{align}\label{equ:likelihood_o}
o_t = X_{t} \ominus X_{t-1} + v, \quad v\sim \mathcal{N}(0,Q),
\end{align}
where $\ominus$ represents an operator that takes two pose and return the relative pose between them in $SE(2)$ or $SE(3)$, and $Q$ is the odometry noise covariance matrix. 
The likelihood of $o_t$ given the two poses is then:
\begin{equation}\label{equ:odom}
p(o_t; X_{t},X_{t-1}) \sim \mathcal{N}(X_t \ominus X_{t-1}, Q)\end{equation}

During navigation, the robot also observes landmarks from the environment.
Assuming that there exist $M$ landmarks in the environment, which might be unknown ahead of time. The positions of the landmarks are denoted as $\mathbf L= \{L_1,\cdots, L_N\}$.
In the 2D case $L_i\in \mathbb{R}^2$, and in the 3D case $L_i\in \mathbb{R}^3$.
At time $t$, the robot obtains $K_t$ landmark measurements, denoted as $\mathbf z_t=\{ z_t^{1}, z_t^{2},\cdots, z_t^{K_t}\}$.
Each measurement is associated to a unique landmark identifiers, the associations are denoted as $\mathbf y_t=\{ y_t^1, y_t^2,\cdots y_t^{K_t}\}$, where $y_t^i \in \{1,\cdots, M\}$. For example, at time 0, the robot obtained two measurements, $z_0 = \{z_0^1,z_0^2\}$. And these 2 measurements are from landmark 5 and 7, then $y_0=\{y_0^1, y_0^2\}=\{5,7\}$. 

Using the standard model that object measurements $z_t^k$ are corrupted by additive Gaussian noise:
\begin{equation}\label{equ:likelihood_z}
z_t^{k} = L_{y_t^k} \ominus X_t + w, \quad w\sim \mathcal{N}(0, R)
\end{equation}
where $R$ is the measurement noise matrix. The likelihood of $z_t^k$ given the robot pose, the landmark association and landmark pose is then:
\begin{align}
p(z_t^k; X_{t},L_{y_t^k}) \sim \mathcal{N}(L_{z_t^k} \ominus X_t, R)
\end{align}

Combining \cref{equ:likelihood_o} and  and \cref{equ:likelihood_z}, the joint log likelihood of odometry and landmark measurements is:
\begin{align}
 & \log p(\mathbf o_{1:T}, \mathbf z_{0:T}; \mathbf X_{0:T}, \mathbf L) \notag \\
 = & \sum_{t=1}^T \phi(o_t; X_{t-1},X_t)+ \sum_{t=0}^T \sum_{k=1}^{K_t} \phi(z_t^{k}; X_t,L_{y_t^k})
\end{align}
where  $\phi(o_t; X_{t-1},X_t)$ and $\phi(z_t^{k}; X_t,L_{y_t^k})$ are odometry and landmark factors respectively.
Using the probability distribution formula for Gaussian noise, it can be shown that each factor follows a quadratic form: 
\begin{align}\label{equ:factors_o_z}
& \phi(o_t; X_{t-1},X_t) \notag \\
= & -\frac{1}{2} \left(X_{t} \ominus X_{t-1}-o_t\right) Q^{-1} \left(X_t \ominus X_{t-1}-o_t\right) \notag \\
& \phi(z_t^{k}; X_t,L_{y_t^k}) \notag \\
=& -\frac{1}{2} \left(L_{y_t^k} \ominus X_t-z_t^{k}\right) R^{-1} \left(L_{y_t^k} \ominus X_t-z_t^{k} \right) 
\end{align}
The pose graph SLAM problem optimizes robot poses $\mathbf X_{0:T}$ and object locations $\mathbf L$ such that the log likelihood is maximized:
\begin{equation}\label{equ:slam_ML}
\max_{\mathbf X_{0:T}, \mathbf L} \log p(\mathbf{o}_{1:T}, \mathbf z_{0:T}; \mathbf X_{0:T}, \mathbf L).
\end{equation}

Note that the log likelihood is nonlinear in $X_t$ and $L_t$ as $\ominus$ is a nonlinear operation in \cref{equ:likelihood_o} and \cref{equ:likelihood_z}.

\begin{figure}[t!]
	\centering
\resizebox{.8\columnwidth}{!}{%	
	\begin{tikzpicture}
	[hidden/.style={circle, draw, minimum size = 0.8cm},
	factor/.style={rectangle, draw, minimum size = 0.2cm}]
% hidden variables
	\node[hidden] (x0) at (-4,0) {$X_0$};
	\node[hidden] (x1) at (-2,0) {$X_1$};
	\node[hidden] (x2) at (0,0) {$X_2$};
	\node[hidden] (x3) at (2,0) {$X_3$};
	\node[hidden] (x4) at (4,0) {$X_4$}; 
	
	\node[hidden] (l1) at (-2.6,2) {$L_1$};
	\node[hidden] (l2) at (-0.4,2) {$L_2$};
	\node[hidden] (l3) at (2,2) {$L_3$};

	\node[factor] (o01) at (-3,-0.5) {$o_{01}$};
	\node[factor] (o12) at (-1,-0.5) {$o_{12}$};
	\node[factor] (o23) at (1,-0.5) {$o_{23}$};
	\node[factor] (o34) at (3,-0.5) {$o_{34}$};

	\node[factor] (z01) at (-3.3,1) {$z_0^1$};
	\node[factor] (z11) at (-2.3,1) {$z_1^1$};
	\node[factor] (z12) at (-1.2,1) {$z_1^2$};
	\node[factor] (z22) at (-0.2,1) {$z_2^1$};
	\node[factor] (z23) at (1,1) {$z_2^2$};
	\node[factor] (z33) at (2,1) {$z_3^1$};
	\node[factor] (z43) at (3,1) {$z_4^1$};
			
	% edges 
	\draw[blue,thick] (x0) --(o01); \ draw[blue,thick] (o01) --(x1);		
	\draw[blue,thick] (x1) --(o12); \draw[blue,thick] (o12) --(x2);
	\draw[blue,thick] (x2) --(o23); \draw[blue,thick] (o23) --(x3);
	\draw[blue,thick] (x2) --(o23); \draw[blue,thick] (o23) --(x3);	
	\draw[blue,thick] (x3) --(o34); \draw[blue,thick] (o34) --(x4);

	\draw[red,thick] (x0) --(z01); \draw[red,thick] (z01) --(l1);		
	\draw[red,thick] (x1) --(z11); \draw[red,thick] (z11) --(l1);		
	\draw[red,thick] (x1) --(z12); \draw[red,thick] (z12) --(l2);
	\draw[red,thick] (x2) --(z22); \draw[red,thick] (z22) --(l2);
	\draw[red,thick] (x2) --(z23); \draw[red,thick] (z23) --(l3);	
	\draw[red,thick] (x3) --(z33); \draw[red,thick] (z33) --(l3);
	\draw[red,thick] (x4) --(z43); \draw[red,thick] (z43) --(l3);	
	\end{tikzpicture}}
	\caption[Pose Graph for SLAM]{Pose Graph for SLAM. $X_t$ denote robot poses, $L_i$ denote landmarks, blue edges denote odometry and red edges denote landmark measurements. }\label{fig:posegraph}
\end{figure}
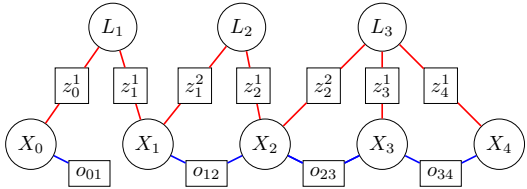

A factor graph representation for SLAM is also referred to as a pose graph. \Cref{fig:posegraph} illustrates such a pose graph. Variables represent either a robot pose or a landmark. And factors are either odometry or landmark measurements.

% % % % % % % % % % % % % % % % % % % % % % % %
\section{Nonparametric Pose Graph}
This section sets up the joint data association and SLAM problem by extending the current pose graph to a novel nonparametric pose graph that tightly couples object association with robot poses.
A new algorithm is also introduced to jointly infer the data association and perform SLAM with this new model.

\subsection{Factor Graph with Multi-class Objects}
Before we move into nonparametric factor graph for imperfect data association, first 
notice in object SLAM, except for measuring the 3D location of objects, we also observe an object class. The observed object class is not always reliable, thus we first establish the probabilistic model for object classes.
Assume there are $N$ object classes in total. For object $i$, denote $u$ as an observation of the object class. The likelihood of $u$ is modeled with a Categorical distribution:
\begin{align}
p(u=j) = \pi_i(j), \quad j=1, \cdots, N
\end{align}
Denote $\pi_i = \{ \pi_i(0), \cdots, \pi_i(N) \}$, $\sum_{n = 0}^{N} \mathbf{\pi}_i(n)= 1$. And if the true object class is $j$, we have $\pi(j) \gg \pi(k)$ for $k\neq j$.

Notice $1 \leq u \leq N$, but we especially design $\pi_i(0)$ to represent the probability of false positives. This design would help the algorithm to filter non-exist object detections in real-world experiments.   

In order to have closed form updates, we apply Dirichlet prior to $\pi_i$ for object $i$:
\begin{align}
\pi_i \sim \text{Dir}(\beta_i).
\end{align}
when there is an observation of class $j$, $u=j$, the posterior distribution of $\pi_i$ is:
\begin{align}
\pi_i |u \sim \text{Dir}(\beta_i+e_{j}).
\end{align}
where $e_j$ represents a unit vector with $j$th element to be 1.

Notice $\beta_i(0)$ represents the initial likelihood of object $i$ to be a false positive. Since observations cannot be 0, when there are more and more observations of object $i$ being obtained, the posterior $\beta_i(0)$ will monotonically decrease. This is consistent with the intuition that if repeated observations are obtained from some object, then it has lower chance to be a false positive.

Combine the multi-class probabilistic setting with the original SLAM problem: each object measurement would be a pair $\{z_t^k, u_t^k\}$, where continuous variable $z_t^k$ represents the 3D location measurement, and discrete variable $u_t^k$ represents the observed object class. 
Recall that $y_t^k =i\in \{1, \cdots, M\}$ represents that the $k$-th measurement at time $t$ is from object $i$. Then $u_t^k$ is a sample from the posterior distribution $\pi_{y_t^k}$.
\begin{align}\label{equ:likelihood_u}
p(u_t^k = j) = \pi_{y_t^k}(j), \quad j\in\{1,\cdots, N\}
\end{align}

The joint log likelihood becomes:
\begin{align}\label{equ:object_u}
& \log p(\mathbf{o}_{1:T}, \mathbf z_{0:T}, \mathbf u_{0:T}; \mathbf X_{0:T}, \mathbf L) \notag \\
 = & \sum_{t=1}^T p(o_t; X_{t-1},X_t)+ \notag \\
& \sum_{t=0}^T \sum_{k=1}^{K_t} \left( p(z_t^{k}; X_t,L_{y_t^k}) + \log \pi_{y_t^k}(u_t^k) \right) \notag \\
 = & \sum_{t=1}^T p(o_t; X_{t-1},X_t)+  \sum_{t=0}^T \sum_{k=1}^{K_t} p(z_t^{k}; X_t,L_{y_t^k}) \notag \\
 &+ \sum_{t=0}^T \sum_{k=1}^{K_t} \log \pi_{y_t^k}(u_t^k)
\end{align}

The new optimization problem is then
\begin{equation}\label{equ:slam_ML_u}
\max_{\mathbf X_{0:T},\mathbf L, \pi} \log p(\mathbf o_{1:T}, \mathbf z_{0:T}, \mathbf u_{0:T}; \mathbf X_{0:T}, \mathbf{L}, \mathbf \pi).
\end{equation}

Compared to \eqref{equ:slam_ML}, the observed data in problem \eqref{equ:slam_ML_u} further includes object class observations  $\mathbf u_{0:T}$, and the variables to be estimated further include the class of objects $\pi$. From \eqref{equ:object_u}, given data association $\mathbf y_{0:T}$, the joint likelihood can be factorized into the sum of likelihood of $\mathbf z_{0:T}$ and $\mathbf o_{0:T}$, and the likelihood of $\mathbf u_{0:T}$. Therefore, the class classes $\mathbf \pi_{0:T}$ is independent of the robot poses $X_{0:T}$ and object positions $\mathbf L$. Optimizing \eqref{equ:slam_ML_u} is equivalent to solving problem \eqref{equ:slam_ML} and computing the object class posterior $\pi$ independently.

\subsection{Nonparametric Pose Graph}
Now we move to the case that the data association $y_t^{k}$ is unknown and must be established.
Deep learning-based algorithms label each object to be of some class, but do not distinguish between different objects of the same class. 
When there are multiple instances of the same object class, such as multiple chairs in a room, possibilities for data association become combinatorial and thus challenging. Instead of relying on a reliable front-end procedure to associate objects, we use a back-end framework to jointly infer the data association and object locations. Note that 
Because of the ambiguous data association, the total number of objects $M$ is unknown ahead of time, and needs to be established as well.
Nonparametric models are a set of tools that adapt the model complexity to data. It has the embedded mechanism that the model parameters could grow when there are new data being observed. In particular, 
Dirichlet Process (DP) is such a nonparametric stochastic process that models discrete distributions but with flexible parameter size. It can be taken as the generalization of a Dirichlet distribution with infinite dimension. Same as Dirichlet distribution is the conjugate prior for a categorical distribution, DP can be viewed as the conjugate prior for infinite, nonparametric discrete distributions \cite{Ferguson1973}. In this work, we use a Dirichlet Process (DP) as the prior for data associations $y_t^k$. In particular, assume at any point, there are $M$ objects being detected in total, the probability of $y_t^k$ belongs to object $i$:
\begin{align}\label{equ:DPprior}
p(y_t^k=i) = \text{DP}(i) = \left\{ \begin{array}{ll}
\frac{m_i}{\Sigma_i m_i+\alpha} & 1\leq i \leq M ,\\
\frac{\alpha}{\Sigma_i m_i+\alpha} & i=M+1.
\end{array}\right.
\end{align}
where $m_i$ is the number of measurements of object $i$, and $\alpha$ is the concentration parameter of DP prior that determines how likely it is to create a new object.
The intuition behind this model is that the probability $y_t^k$ is from some existing object $i\leq M$ is proportional to the number of measurements of object $i$, and the probability $y_t^k$ is from a new object $M+1$ is proportional to $\alpha$.

The joint log likelihood of odometry $\mathbf{o}_{0:T}$, object measurement $\mathbf{z}_{0:T}$ and object classes $\mathbf u_{0:T}$ given data association $\mathbf y_{0:T}$ is
\begin{align}\label{equ:object_mea_z}
 &\log p(\mathbf{o}_{1:T}, \mathbf z_{0:T}, \mathbf u_{0:T}; \mathbf X_{0:T}, \mathbf y_{0:T}, \mathbf L, \mathbf \pi) \notag \\
= &\sum_{t=1}^T  \phi(o_t; X_{t-1},X_t)+ \sum_{t=0}^T\sum_{k=1}^{K_t} \left( \pi_{y_t^k}(u_t^k)+ \phi (z_t^k; X_t, L_{y_t^k})\right).\notag
\end{align}
The joint log likelihood \eqref{equ:object_mea_z} has the same form as \eqref{equ:object_u}. However, in \eqref{equ:object_mea_z}, the likelihood of object measurements $\mathbf z_{0:T}$ and object classes $\mathbf u_{0:T}$ are correlated through data association $\mathbf y_{0:T}$.

The new optimization problem is then over The joint log likelihood of odometry $\mathbf{o}_{0:T}$, object measurement $\mathbf{z}_{0:T}$ and object classes $\mathbf u_{0:T}$ given data association $\mathbf y_{0:T}$:
\begin{equation}\label{equ:slam_ML_z}
\max_{\mathbf X_{0:T},\mathbf L,\mathbf y_{0:T}, \pi} \log p(\mathbf o_{1:T}, \mathbf z_{0:T}, \mathbf u_{0:T}; \mathbf X_{0:T}, \mathbf{L},\mathbf y_{0:T}, \pi).
\end{equation}

Compared with \Cref{equ:slam_ML}, the new optimization problem \Cref{equ:slam_ML_z} is more challenging in that data associations $\mathbf y_{0:T}$ are unknown.
As a result, log probabilities of object measurements %\Cref{equ:object_mea_z} 
no longer have a simple form, and the problem \Cref{equ:slam_ML_z} becomes a mixed integer nonlinear problem.
Secondly, the number of true objects in the environment $M$ is not necessarily known a priori, problem \Cref{equ:slam_ML_z} must infer $M$ at the same time.

\begin{figure}[t!]
	\centering
\resizebox{.75\columnwidth}{!}{%	
	\begin{tikzpicture}
	[hidden/.style={circle, draw, minimum size = 0.8cm}]
	
	% hidden variables
	\node[hidden] (x0) at (0,0) {$X_0$};
	\node[hidden] (x1) at (2,0) {$X_1$};
	\node[hidden] (x2) at (4,0) {$X_2$};
	\node[] (xd) at (6,0) {$\dots$};
	\node[hidden] (xT) at (8,0) {$X_T$};
	
	\node[hidden, thick] (z0) at (0,-1.5) {$y_0$};
	\node[hidden, thick] (z1) at (2,-1.5) {$y_1$};
	\node[hidden, thick] (z2) at (4,-1.5) {$y_2$};
	\node[] (zd) at (6,-1.5) {$\dots$};
	\node[hidden, thick] (zT) at (8,-1.5) {$y_T$};	

	\node[hidden,align=center] (y) at (4,2) { $\mathbf L$, $\pi$};	
	
	% edges 
	\draw[red,thick] (x0)  --(x1);
	\draw[red,thick] (x1)  --(x2);
	\draw[red,thick] (x2)  --(xd);	
	\draw[red,thick] (xd) --(xT);
	
	% meas
	\draw[blue,thick] (x0) --(z0);
	\draw[blue,thick] (x1) --(z1);	
	\draw[blue,thick] (x2) --(z2);	
	\draw[blue,thick] (xT) --(zT);
	
	\draw[blue,thick] (x0) --(y);	
	\draw[blue,thick] (x1) --(y);
	\draw[blue,thick] (x2) --(y);
	\draw[blue,thick] (xT) --(y);			
	\end{tikzpicture}}
	\caption[Factor graph for SLAM with imperfect data association]{Factor graph for SLAM with imperfect data association. $y_t$ represents the data association: the measurement at time $t$ is from object $y_t$. In SLAM with imperfect data association, $y_t$ is unknown and must be established at the same time.}
	\label{fig:factor_slam_z}
\end{figure}
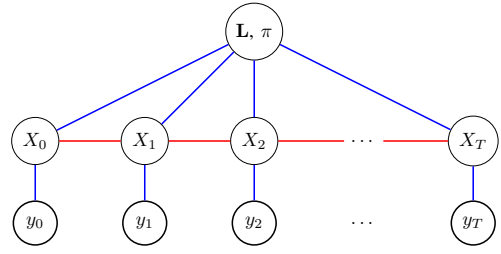

\subsection{Nonparametric SLAM}
From the last section, for $t = 1, \cdots, T, k=1, \cdots, K_t$, the generative model for our problem is
\begin{subequations}
\begin{align}
	y_t^k & \sim \text{DP}(\alpha), \\
	\pi_{y_t^k} & \sim \text{Dir}(\beta_{y_t^k}), \\
	o_t & \sim \mathcal{N}(X_{t}\ominus X_{t-1}, Q), \\ 
	u_t^k & \sim \text{Cat}(\pi_{y_t^k}) , \\
	z_t^k  &\sim \mathcal{N}(L_{y_t^k}\ominus X_t, R),
\end{align} 
\end{subequations}
where $\alpha$, $\beta$, $Q$, and $R$ are given parameters.
Robot poses $\mathbf X_{0:T}$, landmark locations $\mathbf L$, object class distributions $\mathbf \pi_{1:M}$ and object associations $\mathbf y_{0:T}$ are variables to be estimated.
The odometry $\mathbf o_{1:T}$ and object measurements $\mathbf z_{0:T},\mathbf u_{0:T}$ are observed data.

Different from a canonical DP mixture model, the observed data $\mathbf z_{0:T}$, $\mathbf u_{0:T}$, and $\mathbf o_{0:T}$ are not independent samples given variables $\mathbf X_{0:T}$, $\mathbf L$, and $\pi$, but are correlated through the factor graph.
Therefore, the inference involves computing maximum likelihood over factor graphs.
When both associations and variables are to be established, standard approaches alternate between assigning data and optimizing variables.
In the case of known object number $M$, K-means has a deterministic data association, while expectation-maximization associates data in a probabilistic way \cite{Bishop07_PRML}.
When the number of objects is not known a priori and DP is used as prior, Markov Chain Monte Carlo methods (e.g. Gibbs sampling) or variational inference algorithms are widely used \cite{Bishop07_PRML}.
However, in these algorithms, the likelihood of each label $y_t^k$ to be any underlying object $\mathbf L$ needs to be computed and tracked all the time. The algorithm will need to go through all of the data multiple times to converge to a steady state distribution.
The large scale and strong dependence of data in our problem make such approaches inappropriate.

It is shown in \cite{DPmeans} that under the small variance assumptions, Gibbs sampling can be simplified to DPmeans. Instead of sampling the posterior distribution, $y_t$ is assigned to be the maximum likelihood object if the likelihood is within some certain threshold, otherwise it is assigned to a new object.
Intuitively, in this case, small variance means that the noise in odometry, object measurement and object class is relative small, so that the posterior distribution of $y_t$ is peaky.
\begin{assumption}
Variance in odometry, object measurement and object class is small, so that the posterior distribution of data association has small variance and a unique maximal likelihood value.
\end{assumption}

The DPmeans algorithm alternates between two steps: maximize likelihood on variables $\mathbf X_{0:T}, \mathbf L, \pi$, and assign data association $\mathbf y_{0:T}$ to their maximum likelihood objects.
\Cref{alg np-slam} shows the overall flow of the approach. 
And the following explains the algorithm step by step.

\paragraph{Initialization (line \ref{alg:init})} 
In initialization, all $y_t^k$ are set to be an object by its own. Robot poses $\mathbf X_{0:T}$ and object locations $\mathbf L$ are initialized by their open loop estimation. The Dirichlet distribution prior for object class are set to be some initial value $\beta_0$.

\paragraph{Optimizing data association (line \ref{alg:data_association_s})}
 While executing the main loop, the algorithm alternates between optimizing associations $\mathbf y_{0:T}$, and variables $\mathbf X_{0:T}$, $\mathbf L$, and $\mathbf \beta$. When optimizing object association,
  fix $\mathbf X_{0:T}$, $\mathbf L$ and $\beta$, and compute the posterior of $y_t^k$ as the product of its DP prior \eqref{equ:DPprior} and likelihood of measurements $(u_t^k, z_t^k)$ (see \eqref{equ:likelihood_o} and \eqref{equ:likelihood_u}).
 \begin{align}
 p_i \propto \text{DP}(i) p(u_t^k; \pi_i)p(z_t^k; X_t, L_i).
 \end{align}
 Then $y_t^k$ is assigned to the maximum likelihood object
 \begin{align}
 y_t^k = \arg\max_i p_i.
 \end{align}
 
\paragraph{Optimizing poses (line \ref{alg:slam_s})}
When optimizing poses, object associations $y_t^k$ are fixed. The posterior parameters for the Dirichlet distribution of object class can be updated with
\begin{align}
\beta_i(j) \gets \beta_0(j) + \sum_{t,k} \mathbb{I}_{y_t^k=i}\mathbb I_{u_t^k=j},
\end{align}
where $\beta_i$ is the hyper parameter for the Dirichlet prior on $\pi_i$. Notation $\mathbb{I}_{a=b}$ represents indicators whether quantity $a$ equals quantity $b$.
Then $\sum_{k,t} \mathbb{I}_{y_t^k=i}$ is the total number of object detections assigned to object $i$, and $\sum_{k,t} \mathbb{I}_{y_t^k=i}\mathbb{I}_{u_t^k=j}$ represents from the detections of object $i$, how many are class $j$. %\sxx{define the notaiton in more detail}
With Dirichlet prior $\text{Dir}(\beta_i)$, the maximum likelihood(ML) of each object class $i$ is proportional to parameters $\beta_i$:
\begin{align}
\pi_i = \text{ML}(\text{Dir}(\beta_i)).
\end{align}
The maximum likelihood value of robot poses $\mathbf X_{0:T}$ and object locations $\mathbf L$ can then be obtained by standard SLAM solvers (see \eqref{equ:slam_ML}).

\paragraph{Remove false positive (line \ref{alg:false_positive})}
Recall that we set $\pi_i(0)$  to be the probability that object $i$ is a false positive.
In initialization, $\beta_i(0)$ is set to be some positive number.
When new measurements of object $i$ are obtained and accumulated, $\beta_i$ gets updated such that $\beta_i(j)$, $j>0$ becomes bigger compared to $\beta_i(0)$. As a result, $\pi_i(0)$ decrease monotonically. In the last step, we filter out false positives by simply putting a threshold $\epsilon$ on $\pi_i(0)$.

\begin{algorithm}[t]
 \caption{Nonparametric SLAM} \label{alg np-slam}
 \begin{algorithmic}[1]
 \Require {Odometry measurements $o_{1:T}$, Object measurements $\mathbf u_{0:T}$, $\mathbf z_{0:T}$} 
 \Ensure{Poses $\mathbf X_{0:T}$, number of objects $M$, object association $\mathbf y_{0:T}$, object locations and classes $\mathbf L$, $\beta$}
 \State Initialize $\mathbf X_{0:T}$, $\mathbf L$ with open loop prediction, initialize $\beta_i = \beta_0$. Initialize each $y_t^k$ to be an object of its own \label{alg:init}
 \While{not converged}
 \State Fix $\mathbf X_{0:N}$, $\mathbf L$, $\beta$ \label{alg:data_association_s}
\For{Each measurement $y_t^k$}
\State Computer posterior $p_i$ of being object $i$:
\State \quad $p_i \propto \text{DP}(i) p(u_t^k; \pi_i)p(z_t^k; X_t, L_i)$
\State Assign $y_t^k$ to be maximum likelihood association:
\State \quad $y_t^k=\arg\max_i p_i$
\EndFor \label{alg:data_association_e}
\State Fix $\mathbf y_{0:T}$ \label{alg:slam_s}
\For{each object $i$}
\State  update class $\pi$:
\State \quad $\beta_i(j) \gets \beta_i(j) + \sum_{t,k} \mathbb{I}_{y_t^k=i}\mathbb{I}_{u_t^k=j}$
\State \quad $\pi_i = \text{ML}(\text{Dir}(\beta_i))$
\EndFor
\State optimize $\mathbf X_{0:T}$, $\mathbf L$  with standard SLAM solver with \eqref{equ:slam_ML}
 \EndWhile
\State Remove false positive \label{alg:false_positive}
\State \quad $\forall i$, delete object $i$ if $\pi_i(0)>\epsilon$
 \end{algorithmic}
 \end{algorithm}

\section{Experiment}
\subsection{Simulated Dataset}
\begin{figure}[th!]
\centering
\begin{subfigure}[b]{0.48\columnwidth}
	\includegraphics [clip = true, trim=0 0 0 0, height=3cm]{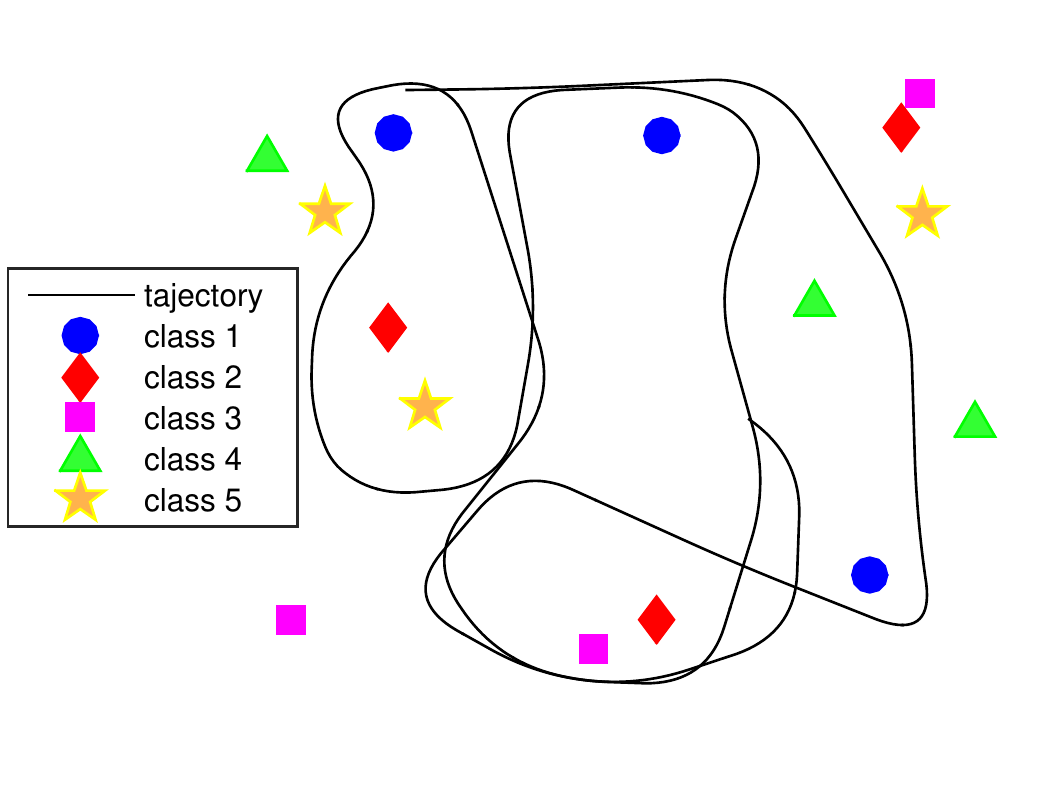}  
	\caption{Ground Truth}\label{fig:sim_truth}
\end{subfigure} 
\begin{subfigure}[b]{0.48\columnwidth}
	\includegraphics [clip = true, trim=20 0 20 0, height=3cm] {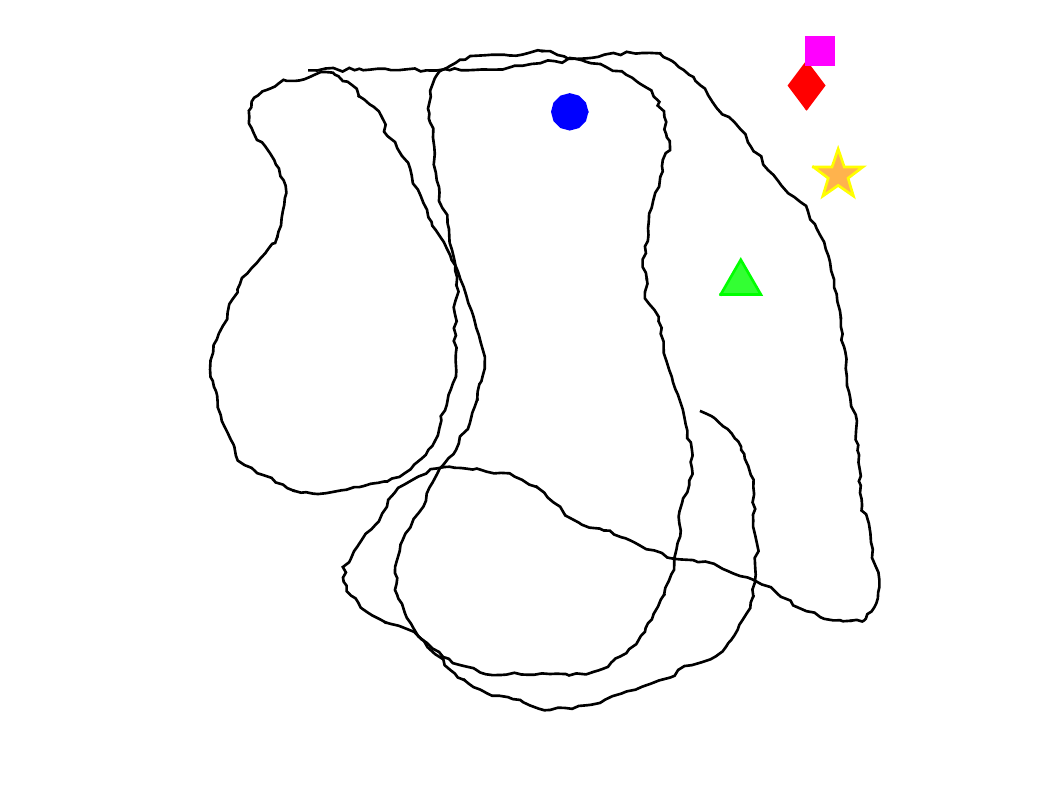}  
	\caption{R-SLAM}\label{fig:sim_NL}
\end{subfigure} 
\begin{subfigure}[b]{0.48\columnwidth}
	\includegraphics [clip = true, trim=20 0 20 0,height=3cm] {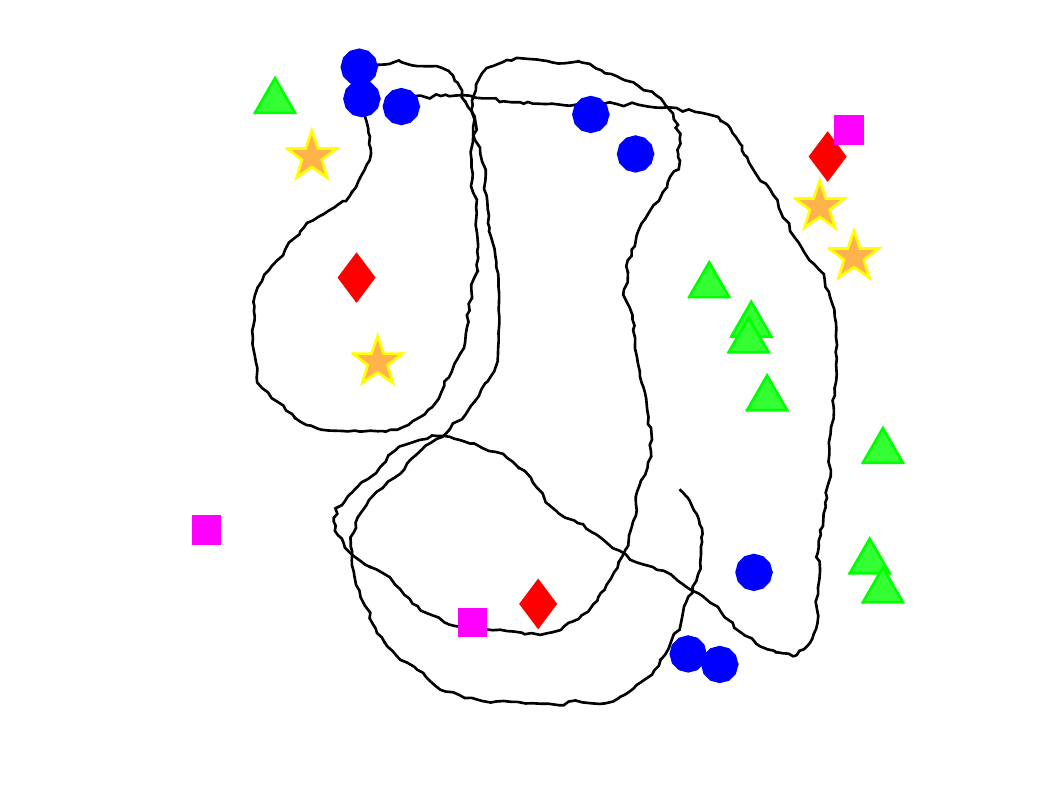}  
	\caption{OL}\label{fig:sim_OL}
\end{subfigure} 
\begin{subfigure}[b]{0.48\columnwidth}
	\includegraphics [clip = true, trim=20 0 20 0,height=3cm] {pr_NP3_15}  
	\caption{NP-Graph}\label{fig:sim_NP}
\end{subfigure} 	\caption{\small Simulation. Black line represents the robot trajectory. Each marker color/shape represent an object class. FbF does neither data associate nor SLAM. OL associate object detection across images but does not optimize robot poses. R-SLAM only uses a subset of consistent object measurements to optimize robot poses. Our approach NP-graph optimizes both robot poses and data association.
}
	\label{fig:sim_solution}
\end{figure}

In the simulation, 15 objects are randomly generated in a 2D plane. They are randomly assigned into 5 different object classes: red diamonds, blue circles, green triangles, yellow stars, and magenta squares. The robot trajectory is manually designed and passes through the environment several times. \Cref{fig:sim_truth} shows the ground truth of the generated dataset. At each pose $X_t$, the robot observes the relative position $o_t^k$ and class $u_t^k$ of the objects that are within its field of view. Gaussian noise are added to the odometry measurements as well as object measurements, see \eqref{equ:likelihood_o}, and \eqref{equ:likelihood_z}. The parameters of the dataset are listed in \cref{tab:sim_dataset}.
\begin{table}[h]
\centering
\caption{\small Simulated Dataset Overview}\label{tab:sim_dataset}
\begin{tabular}{|c|c|}
\hline
\rule{0pt}{8pt}Distance Traveled & 72.7m \\
%\hline
\rule{0pt}{8pt}field of view & 4m, 120 degree \\
%\hline
\rule{0pt}{8pt} no. of odometry measurements & 766 \\
%\hline
\rule{0pt}{8pt}no. of object measurements & 1098\\
%\hline
\rule{0pt}{8pt}odometry noise & $\mathcal{N}(0,0.02^2)$ \\
%\hline
\rule{0pt}{8pt}measurement noise & $\mathcal{N}(0,0.1^2)$\\
\hline
\end{tabular}
\end{table}

\begin{figure}[t]
	\centering
	\begin{subfigure}[b]{0.48\columnwidth}
		\includegraphics [clip = true, trim=0 0 0 0,height=3cm] {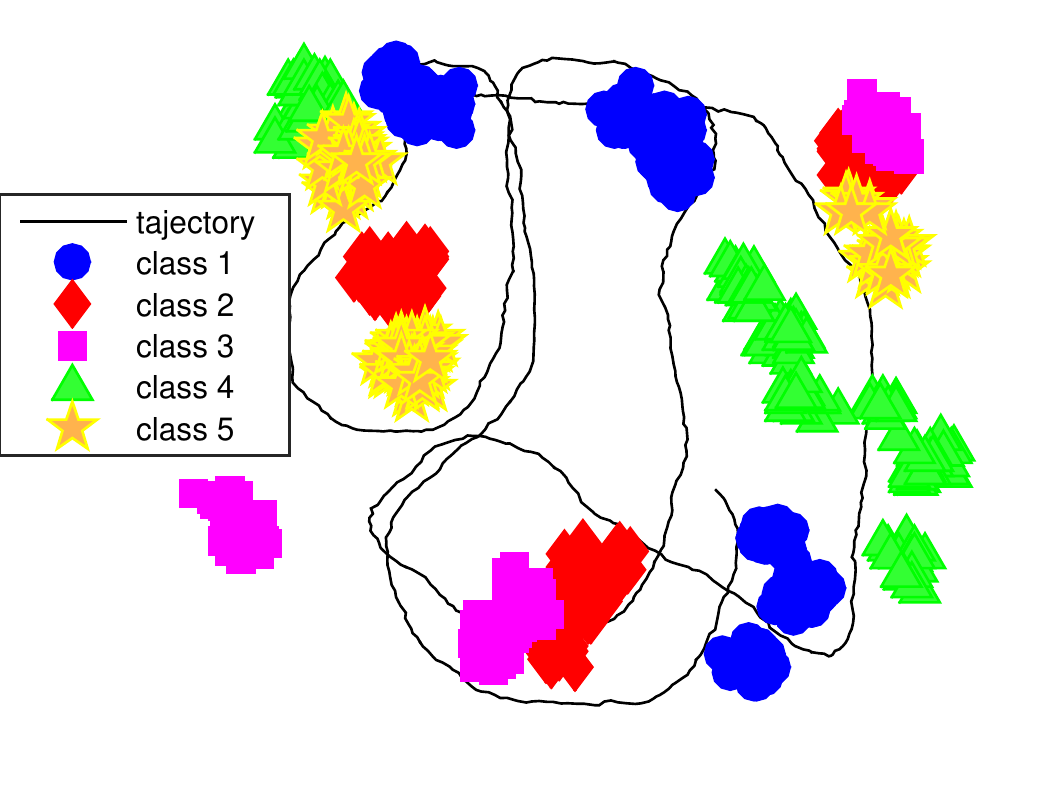}  
		\caption{Iteration 0, 1098 objects}\label{fig:sim_ite0}
	\end{subfigure} 
	\begin{subfigure}[b]{0.48\columnwidth}
		\includegraphics [clip = true, trim=20 0 20 0,height=3cm] {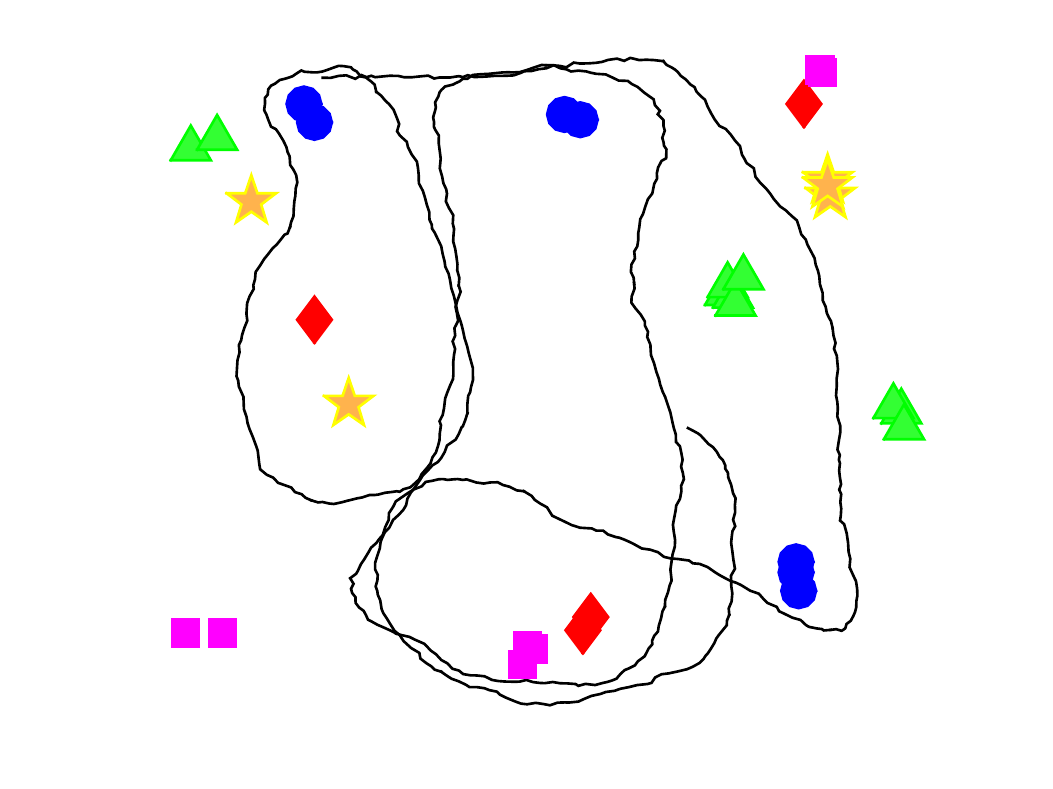}  
		\caption{Iteration 1, 33 objects}\label{fig:sim_ite1}
	\end{subfigure} 
	\begin{subfigure}[b]{0.48\columnwidth}
		\includegraphics [clip = true, trim=20 0 20 0,height=3cm] {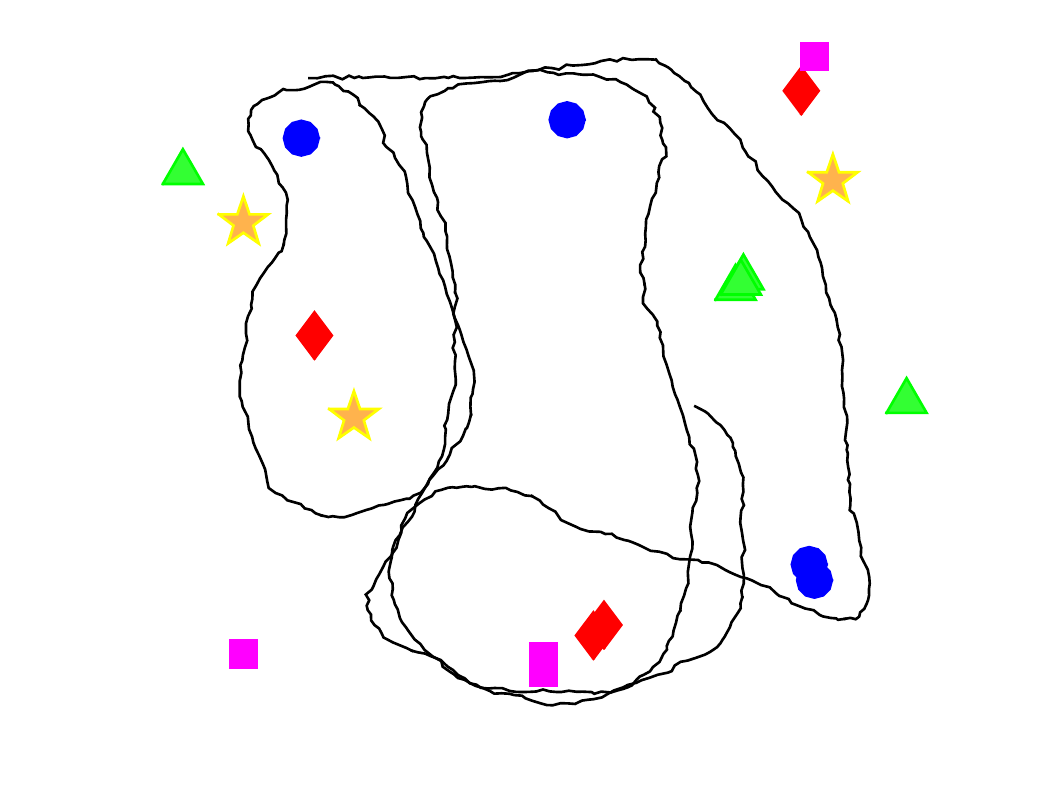}  
		\caption{Iteration 2, 20 objects}\label{fig:sim_ite2}
	\end{subfigure} 
	\begin{subfigure}[b]{0.48\columnwidth}
		\includegraphics [clip = true, trim=20 0 20 0,height=3cm] {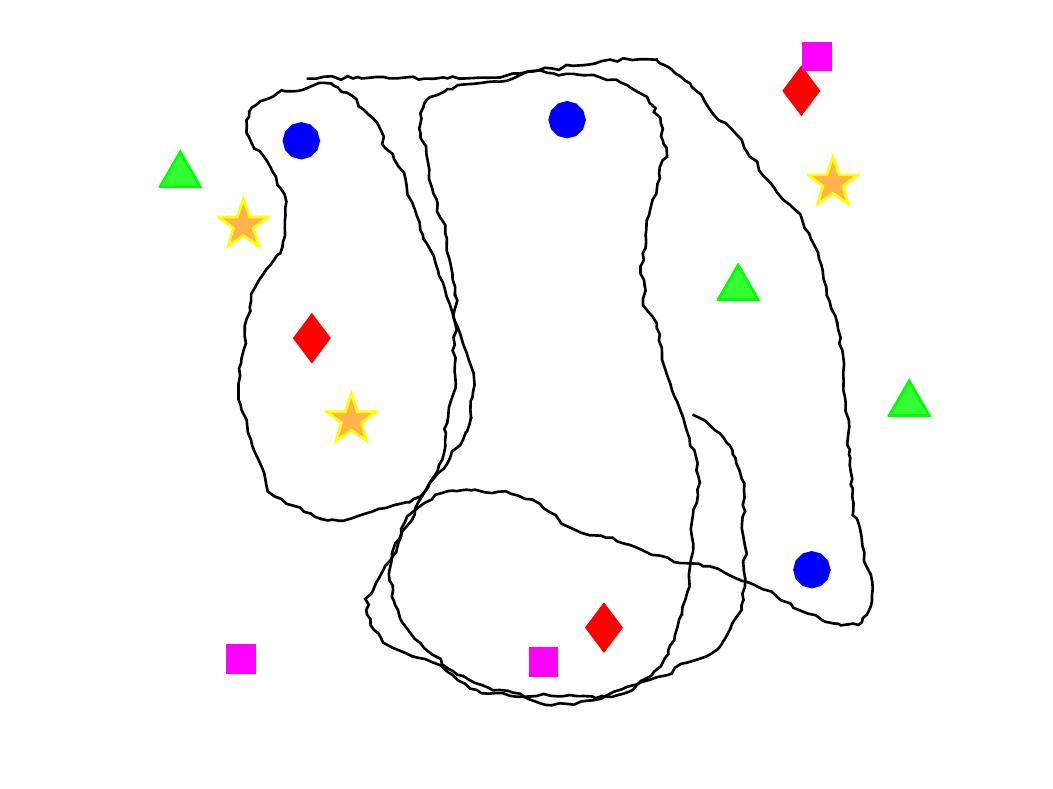} 
		\caption{Iteration 3, 15 objects}\label{fig:sim_ite3} 
	\end{subfigure} 	\caption{Result of nonparametric pose graph at different iterations. Initially there are 1098 object detections. The number reduces to 33 after the first iteration, reduces to 20 after the second iteration, and converges to the ground truth 15 after 3 iterations.} 
	\label{fig:sim_iteration}
\end{figure} 

\Cref{fig:sim_ite0} shows the object predictions based purely on open-loop odometry.
There is significant amount of variance and drift in the distribution of these predicted object locations, which obscures the determination of exactly how many objects there actually are in the environment. The result after the first iteration is shown in figure \ref{fig:sim_ite1}; the nonparametric pose graph clusters the measurements and uses it to correct robot poses.
The total number of objects is reduced to~33. The result after the second iteration is shown in figure \ref{fig:sim_ite2}; the algorithm further reduces the total number of objects to~20.  After three iterations(figure \ref{fig:sim_ite3}), the algorithm converges to the true underlying number of objects, which is~15.

The performance of the proposed nonparametric graph (NP-Graph) is compared to three existing methods:
\begin{enumerate}
\item {\em Frame by frame detection (FbF)}:
each object in each frame is taken as new, and there are neither SLAM nor data association (see \cref{fig:sim_ite0}).
\item {\em Open-loop Object Detection (OL)}:
use robot odometry to perform data association across images, but do not use data association results to correct robot poses (see \cref{fig:sim_OL}).
\item {\em Robust SLAM (R-SLAM)}:
back-end algorithm that finds the maximal set of consistent measurements, but eliminate inconsistent measurements (see \cref{fig:sim_NL}).
\end{enumerate}

\begin{figure}[t]
\centering
	\includegraphics [clip = true, trim=150 280 150 290, width=0.8\columnwidth]{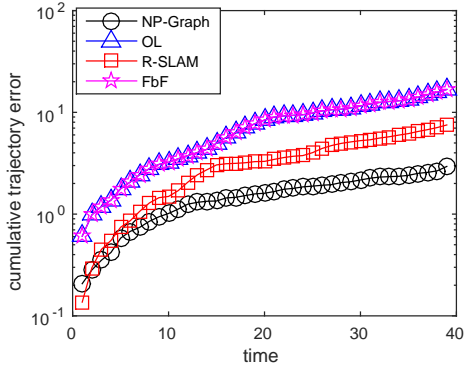}  
	\caption{\small Cumulative robot pose error along the trajectory. NP-graph is able to use noisy label to close loops, thus has magnitude less error than other approaches.}
	\label{fig:sim_err}
\end{figure}

\begin{figure}[h]
\centering
	\includegraphics [clip = true, trim=140 290 150 300, width=0.8\columnwidth]{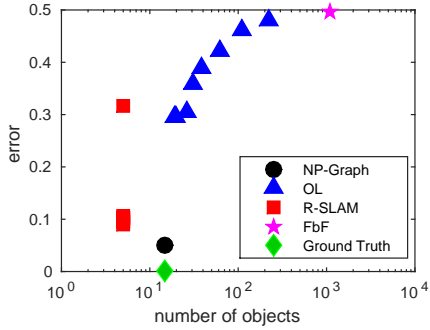}  
	\caption{\small Comparison of number of objects and mean error on objects over multiple-trials. Both FbF and OL have big location error and overestimate number of objects. R-SLAM has much smaller location error, but always underestimate number of objects. NP-graph jointly infers object labels and locations, thus is closest to ground truth in both number of objects and location accuracy.} %\sxx{Too much text in caption. Move to a paragraph.}
	\label{fig:sim_No}
\end{figure}

\Cref{fig:sim_solution} and Table \ref{tab:compare} compare the SLAM performance results of four different algorithms. \Cref{fig:sim_err} shows the cumulative position error of the robot trajectory compared to ground truth, while \Cref{fig:sim_No} compares the number of objects identified and their localization error with a variety of parameter settings for the R-SLAM and OL approaches. 
FbF and OL purely rely on odometry and do not correct robot poses, therefore have the biggest error. R-SLAM uses a subset of object measurements to close loops on robot poses, thus the error is smaller. Our NP-graph based approach make use of all the object measurements, thus has the smallest error on both robot poses and object positions. FbF does not do any data association, thus significantly over estimate the number of objects. The OL approach does not optimize robot poses. When the robot comes back to a visited place, the odometry has drifted significantly thus the OL approach could not associate the objects to the same one observed before. As a result, the OL approach also over estimate the total number of objects. R-SLAM only keeps one set of consistent measurements for each object class, therefore it is only able to detect one instance for each object class, and significantly underestimate the total number of objects. NP-Graph, on the other hand, utilize all of the object measurements and jointly infers both robot poses and the data associations, thus can correctly infer the right number of objects.
\begin{table}[t]
\centering \caption{ Performance Comparison on Simulated Dataset}\label{tab:compare}
\setlength{\tabcolsep}{2pt}\def\arraystretch{.9}%
\begin{tabular}{|c||c|c|c|c|c|} \hline
& mean  & cumulative   & percent of   & number   & mean   \\
&  pose  &  trajectory  & measurements  &  of  &  object  \\
&   error &   error & used &   objects &   error \\
\hline
\rule{0pt}{8pt} NP-Graph & 0.07 & 55.1 & 100 & 15 & 0.05 \\
\rule{0pt}{8pt}OL &  0.42 & 320.6 & 100 & 39 & 0.39 \\
\rule{0pt}{8pt}R-SLAM & 0.20 & 150.5 & 20.2 & 5 &    0.20 \\
\rule{0pt}{8pt}FbF &  0.42 & 320.6 & 100 & 1098 & 0.49\\
\hline
\end{tabular}
\end{table}

%\sxx{Comparision of computation time/complexity of different approaches?}

\subsection{Office Environment}
To test the performance in real-world scenarios, we collected a dataset of an office environment and used deep learning to detect objects, such as chair, screen, cups etc.
The statistics about the office dataset is shown in \Cref{tab:office}.
\begin{table}[h!]
\centering \caption{\small Office Dataset}\label{tab:office}
	\begin{tabular}{|c|c|}
	\hline
	\rule{0pt}{8pt}image resolution & 640$\times$480 \\
	\rule{0pt}{8pt}distance traveled & 28.06m \\
	\rule{0pt}{8pt}during & 167s \\
	\rule{0pt}{8pt}no.\ of odometry & 696 \\
	\rule{0pt}{8pt}no.\ of objects & 30\\
	\rule{0pt}{8pt}no.\ of object detections & 1588 \\
	\rule{0pt}{8pt}odometry noise & $\mathcal{N}(0, 0.1))$ \\
	\rule{0pt}{8pt}measurement noise & $\mathcal{N}(0,0.5)$ \\
	\hline
	\end{tabular}
\end{table}

\begin{figure}[ht]
\centering
\begin{subfigure}[b]{0.48\columnwidth}\centering
	\includegraphics [clip = true, trim=200 250 130 250, height=3cm, angle = 0]{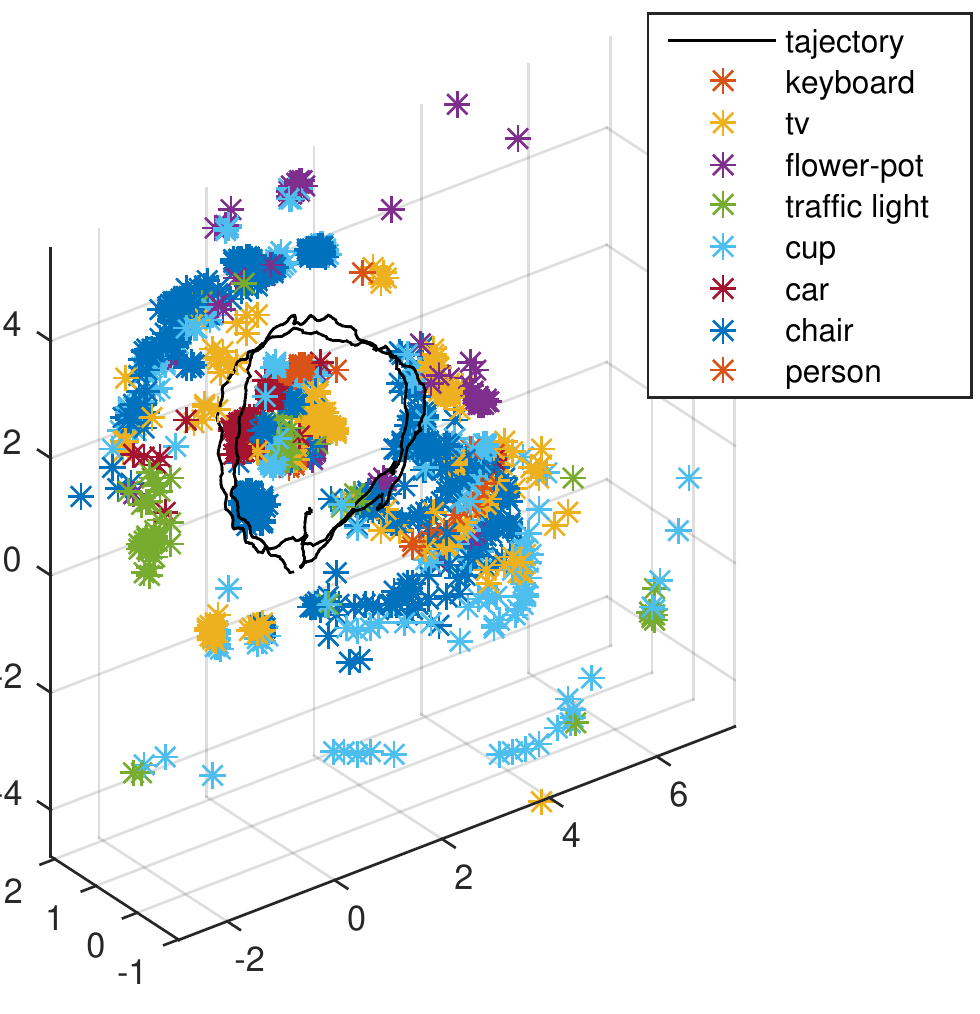}  
	\caption{FbF}\label{fig:office_frame}
\end{subfigure} 
\begin{subfigure}[b]{0.48\columnwidth}\centering
	\includegraphics [clip = true, trim=190 250 100 250, height=3cm, angle = 0]{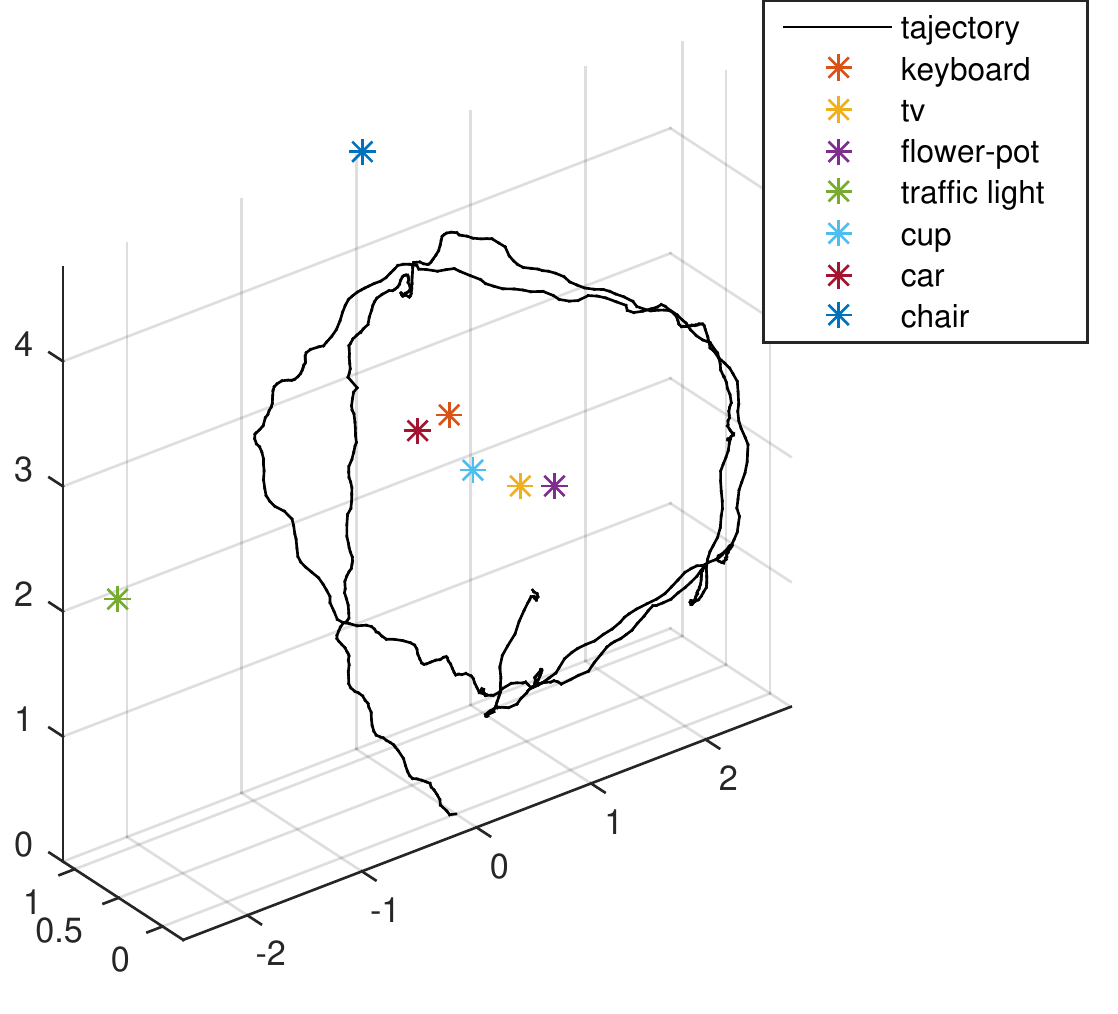}  
	\caption{R-SLAM}\label{fig:office_robust}
\end{subfigure} 
\begin{subfigure}[b]{0.48\columnwidth}\centering
	\includegraphics [clip = true, trim=190 250 150 250, height=3cm, angle = 0]{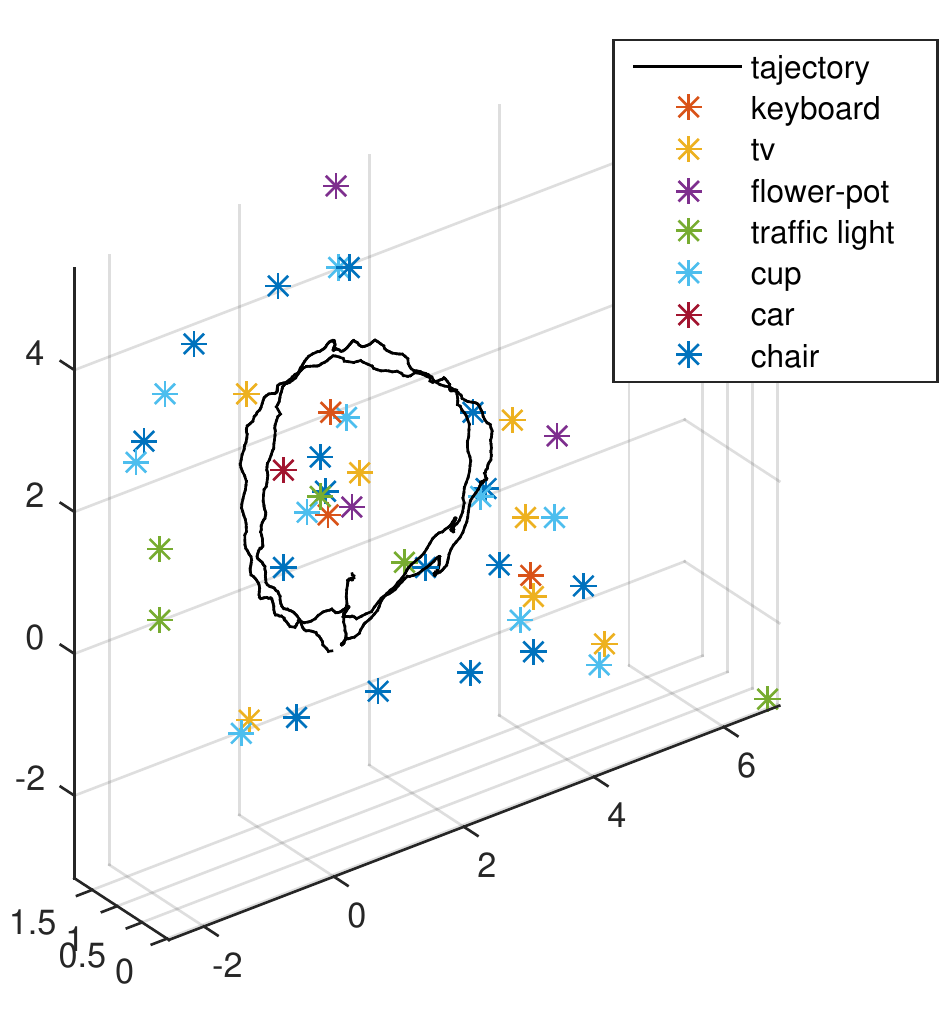}  
	\caption{OL}\label{fig:office_OL}
\end{subfigure} 
\begin{subfigure}[b]{0.48\columnwidth}\centering
	\includegraphics [clip = true, trim=220 250 130 250, height=3cm, angle = 0]{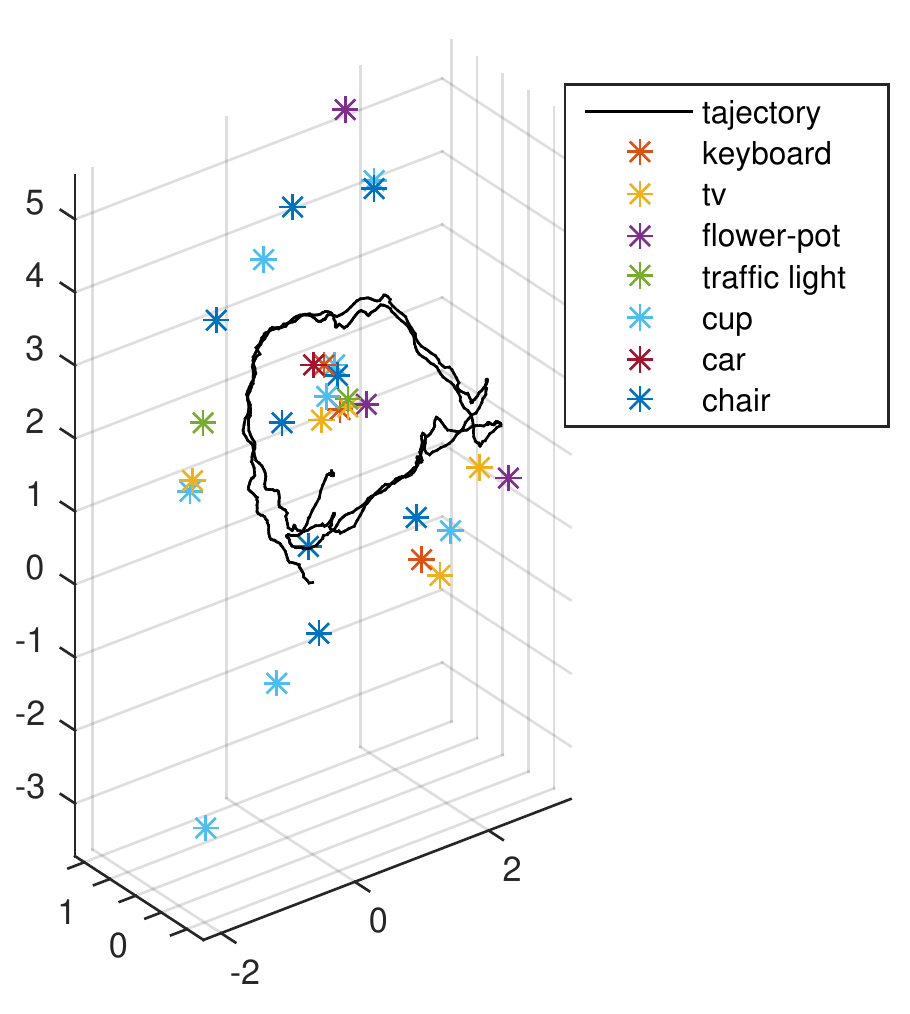}  
	\caption{NP-Graph}\label{fig:office_NP}
\end{subfigure} 
\caption{\small Office Dataset. Black line represent robot trajectory. Markers represent objects. Each color represent an object class. FbF approach does not do data association nor SLAM. R-SLAM does SLAM but not data association. OL approach does data association, but not SLAM. NP-Graph jointly infers data association and does SLAM. It has the least number of objects, data localize the objects, and closes loop thus has least error on robot trajectory.} 
	\label{fig:office_solution}
\end{figure}

\iffalse
The performance of the proposed nonparametric graph (NP-Graph) is compared to three existing methods:
\begin{enumerate}
\item {\em Frame by frame detection (FbF)}:
each object in each frame is taken as new, and there are neither SLAM nor data association (see \cref{fig:office_frame}).
\item {\em Open-loop Object Detection (OL)}:
use robot odometry to perform data association across images, but do not use data association results to correct robot poses (see \cref{fig:office_OL}).
\item {\em Robust SLAM (R-SLAM)}:
back-end algorithm that finds the maximal set of consistent measurements, but eliminate inconsistent measurements (see \cref{fig:office_robust}).
\end{enumerate}
\fi
\Cref{tab:office_compare} and \Cref{fig:office_solution} compare the performance of FbF, R-SLAM, OL and our approach NP-Graph. While the ground truth for object positions is not available for this dataset, we compare the performance on the number of valid objects, the number of inlier measurements and the variance on object positions.
An object is defined as valid when its false positive probability $\pi_i(0)$ is below a threshold ($\epsilon=2\%$), otherwise it is marked as a false positive. A measurement is denoted as an inlier when it is associated with a valid object.
The object variance is determined from the uncertainty in the predicted location of the object from its associated measurements. From \Cref{tab:office_compare}, the NP-Graph has the highest percentage of inlier measurements, the closest number of objects to truth, and the smallest variance on the object locations.

While the ground truth for robot poses is not available, either, we compare the performance qualitatively. 
\Cref{fig:office_floormap} shows the floor map of the environment as well as the robot trajectory. \Cref{fig:office_solution} compares the results of 4 approaches.
FbF and OL estimation are open-loop approaches and over estimate total number of objects.
R-SLAM only uses a subset of the object measurements. It can only identify one instance for each object class, and has bad estimates even it closes loops on robot poses.
On the other hand,  NP-Graph is able to close loops on robot poses and recover the turnings at corners.
While there is no ground truth in the office dataset for computing object localization errors, it is worth noting that there is a sweater hanging on the shelf in the far bottom left corner, our algorithm is able to recover its distance while other approaches failed to.

\begin{figure}[ht!]
\centering
	\includegraphics[height=2cm]{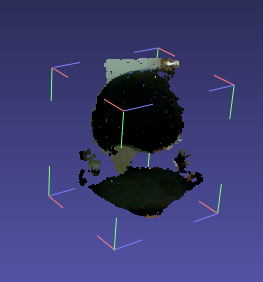}  \hspace{0.2cm}
	\includegraphics[height=2cm]{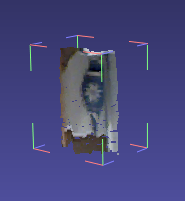} \hspace{0.2cm}
	\includegraphics[height=2cm]{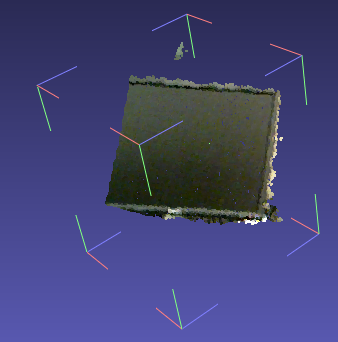} 
	\includegraphics[height=1.6cm]{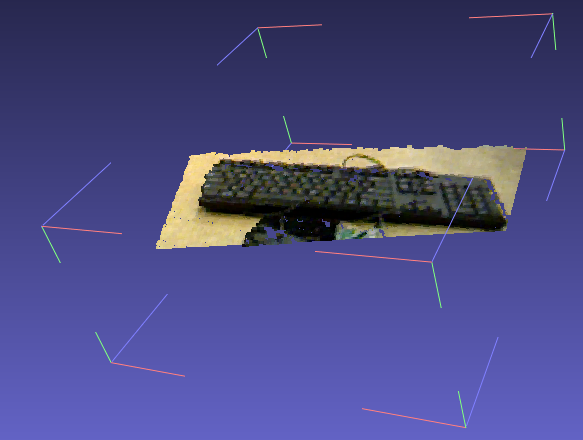} \hspace{0.2cm}
	\includegraphics[height=1.6cm]{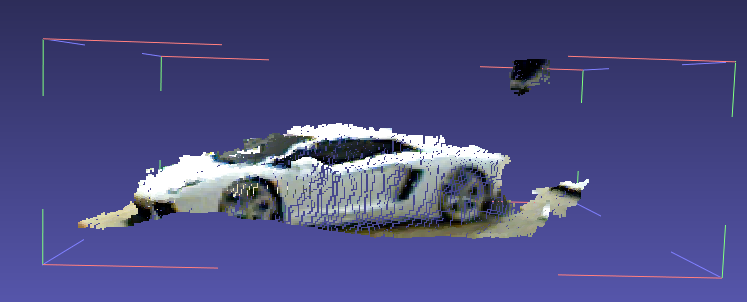}		
	\caption{\small Example of detected objects, plotted from a single frame point cloud. From left to right, top to down are chair, sweater in the corner, screen, keyboard and toy car.} 
	\label{fig:office_eg}
\end{figure}

\Cref{fig:office_eg} shows a few examples of the detected and well associated objects, which includes chair, screen, keyboard, toy car and the sweater hanging in the back corner. These figures are extracted from point cloud of a single bounding box that is associated to the corresponding object. Note that these point clouds are only for illustration purposes, but not maintained in the algorithm. The algorithm only uses the centroid of these point clouds as object measurements.

\begin{table}[t]
\centering \caption{\small Performance Comparison on Office Dataset}\label{tab:office_compare}
\setlength{\tabcolsep}{2pt}\def\arraystretch{.9}%
\begin{tabular}{|c||c|c|c|c|c}\hline
& percentage of & number of  & number of  & variance  \\
& measurement & inlier & false positive  &  on  \\
& inliers & objects &  objects &   objects \\
\hline
\rule{0pt}{8pt} NP-Graph & 88.0 & 31 & 88 & 0.058 \\
\rule{0pt}{8pt}OL &  82.2 & 36 & 175 & 0.121 \\
\rule{0pt}{8pt}R-SLAM & 22.5 & 7 & 0 & 0.225 \\
\rule{0pt}{8pt}FbF &  0 & 0 & 1588 & - \\
\hline
\end{tabular}
\end{table}

\section{conclusion}
Object SLAM is challenging as data association is ambiguous and location measurements unknown. Data association and SLAM are inherently coupled problems. This work proposed a novel nonparametric pose graph that tightly couples these two problems, and developed an algorithm to alternative between inferring data association and performing SLAM. Both simulated and real-world datasets show that our new approach has the capability of doing data association and SLAM simultaneously, and achieves better performance on both associating object detections to unique identifiers and localizing objects.

\section*{ACKNOWLEDGMENTS}
\noindent This research is supported in part by ARO MURI grant W911NF-11-1-0391, ONR grant N00014-11-1-0688 and NSF Award IIS-1318392.

\bibliographystyle{unsrt}
{\small
\bibliography{Beipeng,liam_refs}

\begin{thebibliography}{10}

\bibitem{occupancy_grid}
A.~Elfes.
\newblock Occupancy grids: A stochastic spatial representation for active robot
  perception.
\newblock In {\em Sixth Conference on Uncertainty in AI}, pages 7--24, 1990.

\bibitem{Elfes_1989}
A.~Elfes.
\newblock Using occupancy grids for mobile robot perception and navigation.
\newblock {\em Computer}, 22(6):46--57, june 1989.

\bibitem{Thrun_AR_2003}
S.~Thrun.
\newblock Learning occupancy grids with forward sensor models.
\newblock {\em Autonompous Robots}, 15:111--127, 2003.

\bibitem{prob_robotics}
S.~Thrun, W.~Burgard, and D.~Fox.
\newblock {\em Probabilistic Robotics}.
\newblock The MIT press, Cambridge, Massachusetts, USA, 2005.

\bibitem{Whelan-RSS-15}
T.~Whelan, S.~Leutenegger, R.~S. Moreno, B.~Glocker, and A.~Davison.
\newblock Elasticfusion: Dense slam without a pose graph.
\newblock In {\em Proceedings of Robotics: Science and Systems}, Rome, Italy,
  July 2015.

\bibitem{Keller_dense3d}
M.~Keller, D.~Lefloch, M.~Lambers, S.~Izadi, T.~Weyrich, and A.~Kolb.
\newblock Real-time 3d reconstruction in dynamic scenes using point-based
  fusion.
\newblock In {\em 3D Vision - 3DV 2013, 2013 International Conference on},
  pages 1--8, June 2013.

\bibitem{Newcombe_kinectfusion}
R.~A. Newcombe, S.~Izadi, O.~Hilliges, D.~Molyneaux, D.~Kim, A.~J. Davison,
  P.~Kohli, J.~Shotton, S.~Hodges, and A.~Fitzgibbon.
\newblock Kinectfusion: Real-time dense surface mapping and tracking.
\newblock In {\em IEEE ISMAR}. IEEE, October 2011.

\bibitem{Mahon_IEEETRO_2008}
I.~Mahon, S.B. Williams, O.~Pizarro, and M.~Johnson-Roberson.
\newblock Efficient view-based {{SLAM}} using visual loop closures.
\newblock {\em Robotics, IEEE Transactions on}, 24(5):1002--1014, Oct. 2008.

\bibitem{Graham15_robustslam}
M.C. Graham, J.P. How, and D.E. Gustafson.
\newblock Robust incremental slam with consistency-checking.
\newblock In {\em Intelligent Robots and Systems (IROS), 2015 IEEE/RSJ
  International Conference on}, pages 117--124, Sept 2015.

\bibitem{Rosen12_isam}
D.M. Rosen, M.~Kaess, and J.J. Leonard.
\newblock An incremental trust-region method for robust online sparse
  least-squares estimation.
\newblock In {\em Robotics and Automation (ICRA), 2012 IEEE International
  Conference on}, pages 1262--1269, May 2012.

\bibitem{Kaess_IJRR_2012}
M.~Kaess, H.~Johannsson, R.~Roberts, V.~Ila, J.~J. Leonard, and F.~Dellaert.
\newblock {iSAM}2: Incremental smoothing and mapping using the {B}ayes tree.
\newblock {\em The International Journal of Robotics Research}, 31(2):216--235,
  2012.

\bibitem{montemerlo2003simultaneous}
Michael Montemerlo and Sebastian Thrun.
\newblock Simultaneous localization and mapping with unknown data association
  using fastslam.
\newblock In {\em Robotics and Automation, 2003. Proceedings. ICRA'03. IEEE
  International Conference on}, volume~2, pages 1985--1991. IEEE, 2003.

\bibitem{Graham_robust}
M.C. Graham, J.P. How, and D.E. Gustafson.
\newblock Robust incremental slam with consistency-checking.
\newblock In {\em Intelligent Robots and Systems (IROS), 2015 IEEE/RSJ
  International Conference on}, pages 117--124, Sept 2015.

\bibitem{Everingham10}
M.~Everingham, L.~Van~Gool, C.~K.~I. Williams, J.~Winn, and A.~Zisserman.
\newblock The pascal visual object classes (voc) challenge.
\newblock {\em International Journal of Computer Vision}, 88(2):303--338, June
  2010.

\bibitem{girshick2014rich}
R.~Girshick, J.~Donahue, T.~Darrell, and J.~Malik.
\newblock Rich feature hierarchies for accurate object detection and semantic
  segmentation.
\newblock In {\em Computer Vision and Pattern Recognition (CVPR), 2014 IEEE
  Conference on}, pages 580--587, 2014.

\bibitem{Szegedy_2013_DL}
Christian Szegedy, Alexander Toshev, and Dumitru Erhan.
\newblock Deep neural networks for object detection.
\newblock In C.j.c. Burges, L.~Bottou, M.~Welling, Z.~Ghahramani, and K.q.
  Weinberger, editors, {\em Advances in Neural Information Processing Systems
  26}, pages 2553--2561. 2013.

\bibitem{ILSVRC15}
O.~Russakovsky, J.~Deng, H.~Su, J.~Krause, S.~Satheesh, S.~Ma, Z.~Huang,
  A.~Karpathy, A.~Khosla, M.~Bernstein, A.~C. Berg, and F.~Li.
\newblock {ImageNet Large Scale Visual Recognition Challenge}.
\newblock {\em International Journal of Computer Vision (IJCV)},
  115(3):211--252, 2015.

\bibitem{Erhan_2014_DL}
Dumitru Erhan, Christian Szegedy, Alexander Toshev, and Dragomir Anguelov.
\newblock Scalable object detection using deep neural networks.
\newblock In {\em Proceedings of the 2014 IEEE Conference on Computer Vision
  and Pattern Recognition}, CVPR '14, pages 2155--2162, Washington, DC, USA,
  2014. IEEE Computer Society.

\bibitem{renNIPS15fasterrcnn}
S.~Ren, K.~He, R.~Girshick, and J.~Sun.
\newblock Faster {R-CNN}: Towards real-time object detection with region
  proposal networks.
\newblock In {\em Advances in Neural Information Processing Systems ({NIPS})},
  2015.

\bibitem{Pillai-RSS-15}
S~Pillai and J.~Leonard.
\newblock Monocular slam supported object recognition.
\newblock In {\em Proceedings of Robotics: Science and Systems}, Rome, Italy,
  July 2015.

\bibitem{Song_humanobj}
S.~Song, L.~Zhang, and J.~Xiao.
\newblock Robot in a room: Toward perfect object recognition in closed
  environments.
\newblock {\em CoRR}, abs/1507.02703, 2015.

\bibitem{Atanasov-RSS-14}
N.~Atanasov, M.~Zhu, K.~Daniilidis, and G.~Pappas.
\newblock Semantic localization via the matrix permanent.
\newblock In {\em Proceedings of Robotics: Science and Systems}, Berkeley, USA,
  July 2014.

\bibitem{slam++object}
R.F. Salas-Moreno, R.A. Newcombe, H.~Strasdat, P.H.J. Kelly, and A.J. Davison.
\newblock Slam++: Simultaneous localisation and mapping at the level of
  objects.
\newblock In {\em Computer Vision and Pattern Recognition (CVPR), 2013 IEEE
  Conference on}, pages 1352--1359, June 2013.

\bibitem{Civera_semanticslam}
J.~Civera, D.~Galvez-Lopez, L.~Riazuelo, J.D. Tardos, and J.~M. Montiel.
\newblock Towards semantic slam using a monocular camera.
\newblock In {\em Intelligent Robots and Systems (IROS), 2011 IEEE/RSJ
  International Conference on}, pages 1277--1284, Sept 2011.

\bibitem{Sunderhauf12_robustslam}
N.~Sunderhauf and P.~Protzel.
\newblock Towards a robust back-end for pose graph slam.
\newblock In {\em Robotics and Automation (ICRA), 2012 IEEE International
  Conference on}, pages 1254--1261, May 2012.

\bibitem{Latif13_robust}
Y.~Latif, C.~Cadena, and J.~Neira.
\newblock Robust loop closing over time for pose graph slam.
\newblock {\em The International Journal of Robotics Research}, 2013.

\bibitem{uijlings2013selective}
Jasper~RR Uijlings, Koen~EA van~de Sande, Theo Gevers, and Arnold~WM Smeulders.
\newblock Selective search for object recognition.
\newblock {\em International journal of computer vision}, 104(2):154--171,
  2013.

\bibitem{g2o}
g2o: A general framework for graph optimization.
\newblock \url{https://openslam.org/g2o.html}.

\bibitem{gtsam}
F.~Dellaert.
\newblock Factor graphs and gtsam: A hands-on introduction.
\newblock Technical Report GT-RIM-CP\&R-2012-002, GT RIM, Sept 2012.

\bibitem{Bishop07_PRML}
C.~M. Bishop.
\newblock {\em Pattern Recognition and Machine Learning (Information Science
  and Statistics)}.
\newblock Springer, 1st edition, 2007.

\bibitem{Ferguson1973}
T.~Ferguson.
\newblock {A Bayesian Analysis of Some Nonparametric Problems}.
\newblock {\em The Annals of Statistics}, 1(2):209--230, 1973.

\bibitem{DPmeans}
B.~Kulis and M.~I. Jordan.
\newblock Revisiting k-means: New algorithms via bayesian nonparametrics.
\newblock In {\em Proceedings of the 29th International Conference on Machine
  Learning (ICML)}, pages 513--520, 2012.

\end{thebibliography}
}

\end{document}